\documentclass{article}
\usepackage{enumitem}
\usepackage{multirow}
\usepackage{bm}
\usepackage{fontawesome5}   

\usepackage[preprint]{corl_2026} 

\title{DepthWorld: 3D World Model for Robot Manipulation}

\newcommand{\tinyskip}{\vspace{1pt plus 0.5pt minus 0.5pt}}
\newcommand{\mypar}[1]{\tinyskip\par\noindent\textbf{#1}\nobreak}

\author{
  Jai Bardhan\\
  \texttt{jai.bardhan@cvut.cz} \\
  \And
  Josef Sivic \\
  \texttt{josef.sivic@cvut.cz} \\
  \And
  Vladimir Petrik \\
  \texttt{vladimir.petrik@cvut.cz} \\
  \AND
  \normalfont Czech Institute of Informatics, Robotics and Cybernetics \\
  Czech Technical University in Prague \\ 
  \AND 
  \faGlobe~\url{www.jaibardhan.com/depthworld}
}

\usepackage{graphicx}
\usepackage{amsmath}
\usepackage{amsfonts}
\usepackage{amssymb}
\usepackage{booktabs}
\usepackage[table]{xcolor}
\usepackage{subcaption}
\usepackage{lscape}

\definecolor{consistencyRow}{HTML}{E8F0F8}
\definecolor{groundingRow}{HTML}{FBEFD9}
\definecolor{swapText}{HTML}{8B6508} 

\begin{document}
\addtocontents{toc}{\protect\setcounter{tocdepth}{-1}}

\maketitle



\begin{abstract}
World models offer a data-driven alternative to traditional simulators for robotics, with applications spanning policy evaluation, improvement, and planning. All of these uses depend on faithful 3D geometry, yet current video-based world models are trained on RGB alone and produce rollouts that look correct frame-by-frame but do not compose into a consistent 3D world. Closing this gap requires progress on two fronts: large-scale 3D supervision for manipulation, and an architecture that can absorb it without disturbing strong pretrained video priors. We introduce a calibration pipeline that combines learned stereo depth with a joint factor graph, pooling all episodes collected from the same physical robot to recover its shared kinematic parameters alongside per-scene extrinsics. Applied to the DROID dataset, this yields DROID-3D, a calibrated 3D dataset providing dense metric depth and recalibrated multi-view extrinsics (achieving $<0.7$px reprojection error on 90\% of episodes for external cameras). We then train DepthWorld, a Stable Video Diffusion-based world model that jointly predicts multi-view RGB and depth via spatial latent tiling, leaving the pretrained Variational Autoencoder (VAE) unchanged. Depth supervision improves RGB prediction itself by +1.48 dB PSNR over an identical RGB-only baseline at equal training budget, while simultaneously yielding accurate metric depth for downstream geometric reasoning.

\end{abstract}


\section{Introduction}

World models are a promising data-driven alternative to traditional simulators for robot learning, with the potential to simulate manipulation tasks that are difficult to capture with physics-based engines. Recent work has shown that finetuning pretrained video diffusion backbones with action conditioning on large-scale robotics datasets yields models that generate faithful action-consistent rollouts~\cite{guo2025ctrl,bardhan2026persistent}. Such rollouts are increasingly used for policy evaluation~\cite{guo2025ctrl,bardhan2026persistent}, policy improvement~\cite{guo2025ctrl, unipi}, and planning~\cite{zhen2025tesseract, russell2025gaia}---uses in which the geometric fidelity of the rollout determines whether the signal is usable.

\begin{figure}[t]
    \centering
    \includegraphics[width=\linewidth]{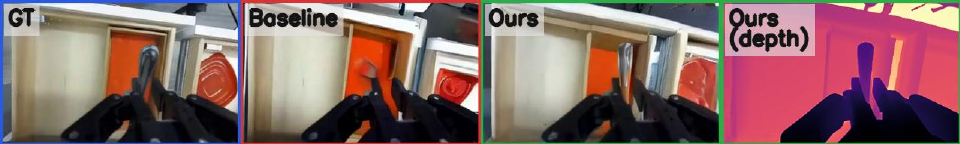}
    \caption{ 
    \textbf{DepthWorld:} The RGB-only baseline fails to maintain the shape of the cutlery, and
the object disappears. DepthWorld capitalizes on the depth inputs to
accurately maintain the cutlery's shape throughout the rollout while
producing high-quality metric depth maps.
    }
    \label{fig:teaser}
    \vspace{-0.5em}
\end{figure}

However, current video world models miss this geometric structure entirely. They are trained on RGB alone, and the training loss provides no signal to resolve the underlying 3D ambiguity of pixel sequences. The resulting models produce visually plausible rollouts whose geometry is internally incoherent: 
predicted depth disagrees across views of the same scene, the model
struggles with geometric reasoning about occlusion and contact
(Figure~\ref{fig:qual-res}), and object shape and identity degrade
over the course of a rollout (Figure~\ref{fig:teaser}).
For downstream uses that read off the rollout, this incoherence is the limiting factor.

The natural fix is to supervise geometry directly during training, the recipe behind recent geometric foundation models~\cite{wang2024dust3r, leroy2024grounding, wang2025vggt, wang2025pi, lin2025depth, keetha2025mapanything}. However, this approach does not directly apply to robot world models for two reasons.
First, geometric foundation models earn their generalization by training on datasets that already carry metric depth and calibrated poses~\cite{yeshwanth2023scannet++,baruch2021arkitscenes,reizenstein2021common,li2018megadepth,yao2020blendedmvs,ramakrishnan2021habitat}. Robot teleoperation datasets such as DROID~\cite{khazatsky2024droid} capture the appearance and dynamics of real manipulation at scale, but their geometric annotations are secondary outputs of the collection pipeline and not at the quality geometric pretraining demands. Lifting them into supervision-quality 3D is non-trivial: manipulation scenes are dominated by textureless and specular surfaces, and even a rig that is internally self-consistent under bundle adjustment can be kinematically decoupled from from the robot's own coordinate frame, severing the critical link to the robot's proprioception and end-effector actions. 
Second, current video world models rely on delicate, pretrained image and video diffusion priors. Conventional architectural modifications for incorporating spatial modalities—such as expanding the input channels of the VAE or introducing parallel dual-branch U-Nets—require destructive weight re-initializations. These interventions risk corrupting the strong visual and motion priors that make these foundation backbones effective in the first place.

We address both through the following contributions:
\begin{enumerate}[leftmargin=1.5em, labelsep=0.4em, nosep]
    \item \textbf{We introduce a calibration pipeline} for recovering metric depth and accurate multi-view extrinsics from any multi-view stereo teleoperation collection with a known URDF. The pipeline combines learned stereo depth with a joint factor graph optimization that pools all episodes of the same physical robot to recover its shared kinematic parameters (joint offsets, hand-eye calibration) alongside per-scene extrinsics. Applied to DROID~\cite{khazatsky2024droid}, this yields \textbf{DROID-3D}---a calibrated 3D corpus providing dense metric depth and recalibrated multi-view extrinsics for over 70{,}000 episodes.
    \item \textbf{We train DepthWorld}, a Stable Video Diffusion-based world model that jointly predicts multi-view RGB and depth via spatial latent tiling that leaves the pretrained VAE unchanged.
    \item \textbf{We show that depth supervision improves RGB prediction}: at equal training budget, DepthWorld gains $+1.48$\,dB PSNR over an RGB-only baseline of identical architecture and data.
\end{enumerate}

\section{Related Work}
\label{sec:related_works}

\mypar{Robotic World Models.}
Action-conditioned video models have emerged as a learned substitute for physics-based simulators in manipulation, with applications spanning policy evaluation~\cite{guo2025ctrl,bardhan2026persistent}, policy improvement~\cite{guo2025ctrl,unipi}, and visual planning~\cite{zhen2025tesseract, russell2025gaia}. Recent works finetune pretrained video diffusion backbones with proprioceptive action conditioning: Ctrl-World~\cite{guo2025ctrl}, an SVD-based controllable world model trained on DROID~\cite{khazatsky2024droid} forms our backbone; IRASim~\cite{zhu2025irasim} along with other works~\cite{jiang2026robowm} systematize the policy-evaluation use case. All these uses presuppose geometrically faithful rollouts.
Current video world models are trained on RGB alone, which does not generally serve as a strong geometric prior.
We close this gap by jointly predicting depth and supervising with point maps.

\mypar{Geometric Foundation Models.}
A recent line of work regresses dense metric geometry directly from images: DUSt3R~\cite{wang2024dust3r} and MASt3R~\cite{leroy2024grounding} predict point maps from image pairs, VGGT~\cite{wang2025vggt} scales this to many-view inputs through 
intermediate camera and DPT~\cite{ranftl2021vision} heads, and MapAnything~\cite{keetha2025mapanything} unifies a broader class of supervision targets. These models inherit their generalization from large datasets~\cite{yeshwanth2023scannet++,baruch2021arkitscenes,reizenstein2021common,li2018megadepth,yao2020blendedmvs,ramakrishnan2021habitat, realestate10k} carrying dense metric depth and calibrated poses, drawn from structured light, LiDAR, and offline reconstruction. 
None of these covers the visual distribution of real manipulation, with its indoor setting and extreme viewpoint differences between fixed external and wrist-mounted cameras. DROID-3D is our contribution at this distribution and scale. 

\mypar{Depth-Aware Video Diffusion.}
Stable Video Diffusion (SVD)~\cite{blattmann2023stable} provides strong appearance and motion priors but is trained on RGB alone. Three approaches have been explored to add depth into this prior. Marigold~\cite{marigold} showed that an image diffusion VAE encodes depth maps essentially without loss,
repurposing the model as a depth predictor; 
DepthCrafter~\cite{hu2025depthcrafter} extends this to video. Joint RGB--depth prediction has been pursued through channel expansion~\cite{zhen2025tesseract, zhang2025imowm}, which adds depth channels to the VAE's input and output convolutions
and through dual-branch architectures~\cite{pallotta2025syncvp} that run a parallel U-Net for depth with cross-connections to the RGB branch. Both modify the pretrained backbone substantially, preventing weight transfer. Our spatial latent tiling extends Marigold's observation
from single-image depth prediction to multi-view RGB--depth video prediction.

\mypar{Robot Datasets and Their Calibration.}
Large-scale teleoperation datasets have grown rapidly in size and diversity---BridgeData~V2~\cite{walke2023bridgedata}, RT-1~\cite{brohan2022rt} and the Open X-Embodiment aggregation~\cite{o2024open}, RH20T~\cite{fang2023rh20t}, and DROID~\cite{khazatsky2024droid}---however; almost all carry geometry as a byproduct of the capture pipeline: monocular RGB in Open~X, single RGB-D in Bridge, and ZED stereo in DROID. Simulation-based corpora such as RoboCasa~\cite{nasiriany2024robocasa} and ARNOLD~\cite{gong2023arnold} carry perfect geometry but inherit the sim-to-real gap that motivates training world models on real data in the first place. DROID is unique among real-world manipulation datasets at this scale in providing synchronized multi-view stereo with known baselines and a publicly released URDF, but its shipped geometric metadata is not at the quality level downstream geometric learning depends on. Post-hoc recalibration of DROID has been attempted: PointWorld~\cite{huang2026pointworld} initializes per-scene extrinsics from VGGT and refines them against robot depth observations, and serves as our baseline in §\ref{sec:eval}.

\section{Constructing Large-scale 3D Datasets for Robot Manipulation}
\label{sec:dataset}

\begin{figure}
    \centering
    \includegraphics[width=\linewidth]{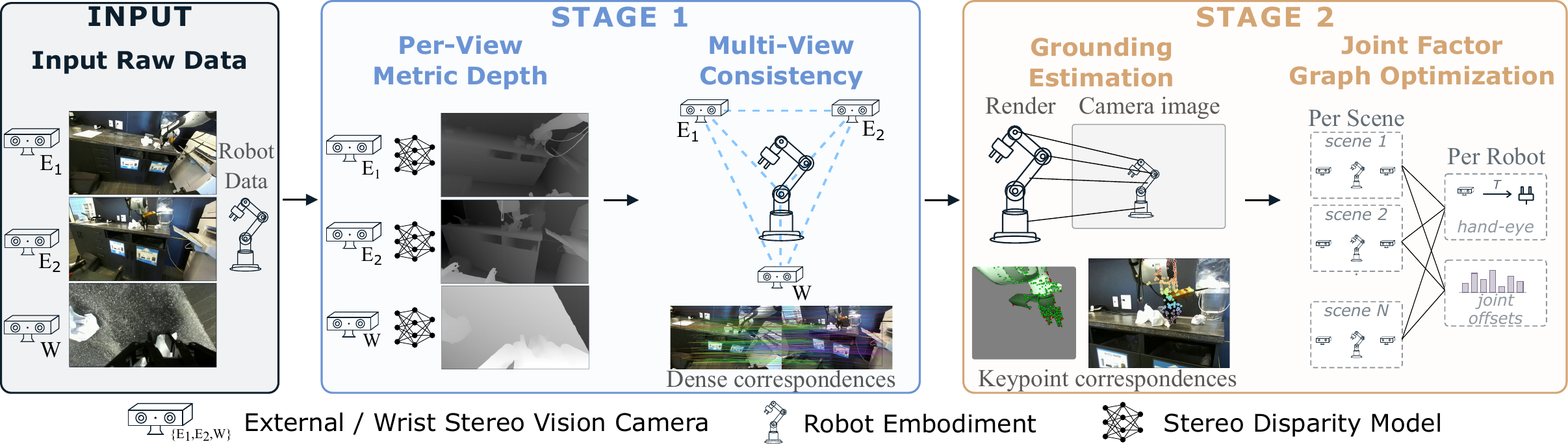}
    \caption{\textbf{DROID-3D calibration pipeline.} Given raw stereo from two external cameras and a wrist-mounted camera together with the robot's URDF and joint encoder readings, \textbf{Stage 1} recovers per-view metric depth from each stereo pair via a learned global-matching network and uses dense correspondences to enforce multi-view consistency. \textbf{Stage 2} renders the URDF into each external view to ground the rig to the robot frame, then forms a joint factor graph that pools all episodes of a robot to recover per-scene extrinsic corrections ($\delta T_{\text{ext1}}^s$, $\delta T_{\text{ext2}}^s$) alongside shared kinematic parameters (hand-eye correction $\delta T_{\text{wrist}}$, joint encoder offsets $\delta \bm{q}$).}
    \label{fig:3d_droid_pipeline}
    \vspace{-0.5em}
\end{figure}

The recent progress in learning-based geometric vision models rests heavily on large-scale datasets providing dense metric depth and calibrated camera poses. The manipulation domain lacks these resources at a comparable scale. While large teleoperation datasets like DROID~\cite{khazatsky2024droid} provide the necessary visual diversity and embodied dynamics, their raw geometric annotations are not at the supervision quality required for geometric pretraining.

We propose an approach that takes a multi-view stereo teleoperation dataset with a known kinematic model (URDF) as input and produces a calibrated RGB-D corpus in the robot’s coordinate frame. Our approach addresses the domain's unique visual and calibration challenges in two distinct stages, shown in Figure~\ref{fig:3d_droid_pipeline} and described in the following sections.

\subsection{Stage 1: Initial Multi-View Calibration}

Stage 1 turns the independently calibrated cameras of each scene into an internally consistent rig placed approximately in the robot's coordinate frame, good enough that the robot's URDF can be rendered into each view and matched against the real image in Stage 2.

\mypar{Per-View Metric Depth.}
Each camera is a calibrated stereo pair with a known baseline $b$ and focal length $f$, so metric depth reduces to predicting per-pixel disparity $d$, with $z = fb/d$. Manipulation scenes are dominated by textureless tabletops and specular robot arms, causing classical stereo SDKs to fail. To overcome this visual ambiguity, we recover per-view metric depth from each stereo pair using a learned global-matching network, $S^2M^2$~\cite{min2025s2m2}. This provides the boundary-sharp metric signal necessary to turn cross-view pixel correspondences into reliable 3D constraints (see Appendix~\ref{app:depth} for an ablation against monocular and SDK baselines).

\mypar{Multi-View Consistency.}
The initial camera calibrations lack full multi-view consistency, and the extreme viewpoint differences between external and wrist-mounted cameras defeat sparse keypoint matchers. We utilize the metric depth above alongside dense feature matching via RoMa v2~\cite{edstedt2025roma} to perform robust bundle adjustment.
Because wrist--external matches are markedly noisier than external--external ones, a joint solve over all three cameras settles in poor local minima, so we optimize in two passes. Pass 1 aligns the two external cameras ($E_1, E_2$) into a consistent pair by minimizing their mutual Cauchy photometric residual. Pass 2 places this pair in the robot frame: the wrist camera pose is held fixed from forward kinematics and the factory hand-eye prior, and a single global $SE(3)$ transform of the external pair is solved to best explain the wrist--external correspondences. The output is a per-scene rig that is only approximately in the robot's base frame, since it still inherits the systematic hand-eye and joint-encoder errors of the factory priors; removing these is the role of Stage 2.
Details of the two-pass formulation, correspondence filtering, and sparse-matcher failure modes are in Appendix~\ref{app:consistency}.

\subsection{Stage 2: Robot-Grounded Joint Optimization}

The input to this stage comprises the visually consistent per-scene multi-camera rigs produced in Stage 1, alongside the robot's URDF and raw joint encoder readings. The output is a set of refined camera extrinsics firmly anchored to the robot's physical base frame, coupled with globally optimized, robot-specific kinematic parameters (hand-eye transform and joint encoder offsets).

Achieving this physical grounding presents two primary difficulties. First, visual reprojection consistency is invariant to rigid drift. A camera rig can be internally self-consistent but physically misaligned with the robot's coordinate frame, which breaks the object-arm spatial relationships required for manipulation. Second, it is notoriously difficult to disentangle scene-specific camera calibration noise from systematic, robot-wide errors, such as constant joint encoder biases or gradual hand-eye mount drift. Optimizing scenes independently simply absorbs these hidden, systematic errors into the per-scene extrinsics.

We overcome these challenges through a joint factor graph optimization that pools visual and kinematic signals across all episodes of a given robot. By rendering the robot arm into the camera views using the URDF and matching it to the real images, we tie the extrinsics to the real world. By optimizing all scenes simultaneously, we cleanly separate transient, scene-specific extrinsic noise from the shared kinematic parameters, multiplying the correction signal for the latter by the number of episodes. Details are given next.

\mypar{Kinematic Anchoring via Rendering.}
To supply the absolute anchor to the robot frame that cross-view consistency alone cannot provide (the first difficulty above), we render the robot arm into each external view utilizing the URDF, the corrected joint angles, and the Stage 1 extrinsic estimates, and use LoMA-G~\cite{nordstrom2026loma} to extract 2D-3D correspondences between the rendered template and the real image. Each match links a rendered surface point, whose 3D position in the robot frame is known from the URDF and forward kinematics, to a real-image pixel, yielding a robot-frame ground-truth signal independent of any cross-view photometric cue. Simultaneously, patch-level features~\cite{oquab2023dinov2} extracted by LoMA-G provide a featuremetric alignment term that yields sub-pixel signals in textureless regions. Rendering details (cropping, non-PBR shading) are given in Appendix~\ref{app:anchoring}.

\mypar{Joint Factor Graph Construction.}
We build a single factor graph that spans all $N_s$ episodes of a given robot (indexed $s \in \{1, \dots, N_s\}$). Two kinds of parameters enter the graph: \emph{per-episode} corrections to the external camera poses, which absorb scene-specific calibration noise, and \emph{shared} kinematic parameters of the robot itself, which are tied across every episode of the robot. Concretely, each episode contributes a 6-DoF $SE(3)$ correction $\delta T_{\text{ext1}}^s$ and $\delta T_{\text{ext2}}^s$ for each external camera (12~DoF per episode), and all episodes share a 6-DoF hand-eye correction $\delta T_{\text{wrist}}$ to the CAD-specified gripper-to-wrist-camera mount, together with 7-DoF joint encoder offsets $\delta \bm{q}$.
The wrist camera is rigidly mounted to the gripper, so its pose has no per-episode degrees of freedom — at frame $t$ it is fully determined by the kinematic chain:
\begin{equation}
\label{eq:chain}
  T_{\text{world},\text{wrist}}(t)
  = T_{\text{world},\text{gripper}}\bigl(\bm{q}(t) + \delta \bm{q}\bigr)
  \cdot T_{\text{gripper},\text{wrist}}^{\text{CAD}}
  \cdot \delta T_{\text{wrist}},
\end{equation}
where $T_{\text{world},\text{gripper}}(\bm{q})$ is the base-to-gripper forward kinematics evaluated at the corrected joint angles, $T_{\text{gripper},\text{wrist}}^{\text{CAD}}$ is the CAD-specified gripper-to-wrist-camera mount, and $\delta T_{\text{wrist}}$ is the optimized correction to that mount.

\mypar{Robust Objective.}
The objective unifies these constraints into a single robust loss over the tens to ${\sim}100{,}000$ parameters of one robot's graph (varying with episode count):
\begin{equation}
\label{eq:objective}
\mathcal{L} = \sum_{k \in \{\text{2d},\, \text{3d},\, \text{l},\, \text{ee},\, \text{f}\}} \lambda_k \sum_{s, i} \rho\!\left(\|r_{s,i}^k\|^2;\, c_k\right) + \lambda_q \|\delta \bm{q}\|^2,
\end{equation}
where $\rho(\,\cdot\,;\, c)$ is the Cauchy loss with scale $c$, $s$ indexes episodes, and $i$ indexes residuals within an episode. The five data terms each constrain a different part of the graph: Mahalanobis-whitened wrist--external 2D reprojections ($\text{2d}$) and 3D metric lifting residuals ($\text{3d}$) tie the kinematic chain of Eq.~\eqref{eq:chain}, and thus $\delta T_{\text{wrist}}$ and $\delta \bm{q}$, to the external cameras; URDF-rendered robot reprojections ($\text{l}$) ground the external extrinsics to the robot frame; cross-camera constraints ($\text{ee}$) keep the external pair consistent; and featuremetric alignment ($\text{f}$) adds sub-pixel signal in textureless regions. The Tikhonov term $\lambda_q \|\delta \bm{q}\|^2$ regularizes under-constrained components of $\delta \bm{q}$, letting the more flexible per-episode parameters explain residuals in those directions.

\mypar{Scalable Optimization.}
Solving this massive graph is bottlenecked by scale, but the Hessian exhibits a highly sparse block structure. We leverage a GPU-accelerated Levenberg-Marquardt solver using the Schur complement. This effectively eliminates the per-scene blocks, reducing the bottleneck to a highly efficient $13 \times 13$ global solve per iteration. This drops the computational cost from cubic-in-$N_s$ to $O(N_s \cdot 12^3 + 13^3)$, keeping typical wall time to roughly 12.5 minutes per robot group.

\subsection{DROID-3D} We run this pipeline on DROID's raw
release~\cite{khazatsky2024droid}, the 71k-trajectory split that ships with the
factory camera parameters. The recordings come from 13 institutions
on Franka Panda arms, each with two table-mounted ZED~2 external
cameras and a wrist-mounted ZED~Mini. After per-robot filtering for
broken metadata or unrecoverable initialization, we run across
\textbf{28 robots and 13 labs covering 71{,}100 scenes}. Calibration
quality against a per-scene baseline is reported in
§\ref{sec:eval}; hyperparameters and solver wall-time in Appendix~\ref{app:hparams}.

\section{Building a 3D-Consistent World Model}
\label{sec:wm}

\begin{figure}
    \centering
    \includegraphics[width=\linewidth]{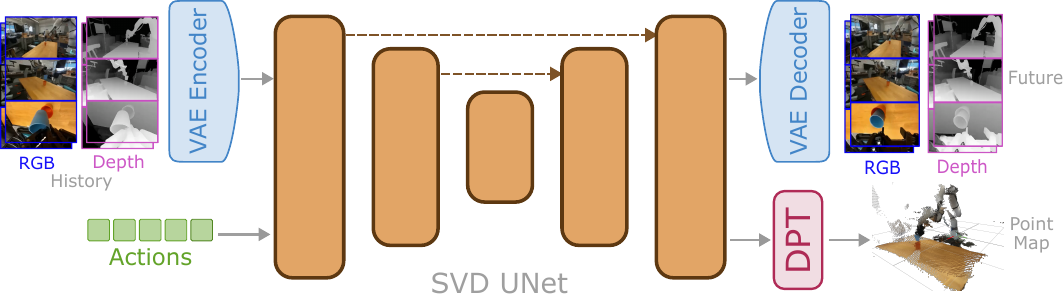}
    \caption{\textbf{DepthWorld Architecture.} Multi-view RGB and
    depth are independently encoded into a spatially tiled latent
    grid ($72 \times 80$) for joint U-Net denoising, preserving
    pretrained SVD priors. Predicted future latents ($x_{0,f}$)
    pass through a DPT head to predict a 3D point map ($X, Y, Z$)
    and confidence logit in the robot base frame, supervised by a
    robust geometric loss.}
    \label{fig:placeholder}
    \vspace{-0.5em}
\end{figure}

The objective of DepthWorld is to generate video rollouts where
predicted depth and RGB compose into a single, geometrically
consistent 3D world, rather than just independent, visually
plausible frames. We achieve this by consuming the dense metric
depth and calibrated extrinsics of DROID-3D as a direct training
signal.

Integrating geometric prediction into state-of-the-art video world
models presents two primary difficulties. First, modifying a
pretrained backbone like Stable Video Diffusion (SVD) (e.g., via
channel expansion or dual-branch networks) destroys the strong
appearance and motion priors that make the model useful in the
first place.  Second, supervising depth independently per-view does not directly encourage 3D consistency across different cameras.

We overcome these challenges through two mechanisms. First, rather
than altering the U-Net architecture, we introduce spatial latent
tiling (§\ref{sec:rgbdpred}) to exploit the Variational Autoencoder's (VAE) natural ability
to encode depth as an image, placing RGB and depth side-by-side in
a wider latent grid. Second, to enhance cross-view consistency, we
apply robot-frame auxiliary supervision (§\ref{sec:auxgeo})
through a lightweight point-map head that projects predictions into
the robot's physical base frame and supervises them against unified
3D point maps. 

\subsection{Joint RGB-Depth Prediction via Spatial Tiling}
\label{sec:rgbdpred}

Marigold~\cite{marigold} showed that an image-diffusion VAE encodes and decodes depth maps faithfully. Building on this, we encode depth as a separate image alongside RGB and tile the two in latent space: each is encoded by the unmodified SVD VAE and placed side-by-side, so the U-Net sees a wider latent grid covering both modalities. The only network change is extending the spatial position embeddings to cover the wider grid.

At each timestep, the model processes images of resolution $192 \times 320$ for $V=3$ cameras (two external, one wrist)
and two modalities (RGB, depth). 
Since the SVD VAE downsamples
spatially by $8\times$, each view-modality tile is independently
encoded into a latent of dimensions $H_{lat} \times W_{lat}$ (where
$H_{lat}=24, W_{lat}=40$). 
We arrange the six tiles in a 2D grid:
RGB and depth tiles for each view are concatenated horizontally
(yielding width $2 \cdot W_{lat}$), and the $V$ views are stacked
vertically (yielding height $V \cdot H_{lat}$). This produces a
single joint latent of shape $72 \times 80$ per timestep (see
Figure~\ref{fig:placeholder}). The U-Net denoises this full grid
as one tensor across a temporal window of 11 frames (6 history, 5
future) at 5~Hz. (See Appendix~\ref{app:depth_preproc} for the depth preprocessing
pipeline required to conform the raw depth to the VAE
distribution.)

\subsection{Robot-Frame Point-Map Supervision}
\label{sec:auxgeo}

Per-view depth denoising trains each predicted depth map against its own target independently, with no explicit penalty when predicted depths disagree across different camera views of the same scene. 
We make cross-view coherence explicit by lifting predictions into the robot base frame and computing the loss there.

We do so by attaching a lightweight DPT~\cite{ranftl2021vision} point-map head
(initialized from VGGT~\cite{wang2025vggt}) to the SVD U-Net's
final predicted latent $\hat{x}_0$. 
For the predicted future
frames $\hat{x}_{0,f}$, we pass the (RGB, depth) latent pairs to
the DPT head, which predicts a 3D point map in robot base coordinates
for each view. 
The ground-truth point map $P_{v,t}^* \in
\mathbb{R}^{H \times W \times 3}$ for view $v$ at frame $t$ is
obtained by backprojecting the GT depth through the camera
intrinsics and transforming it to the robot base frame via the
calibrated extrinsics from §\ref{sec:dataset}.

The head outputs four channels per frame and view: spatial
coordinates $X, Y, Z$ (the point-map prediction $P_{v,t}$ in robot
world frame) and a per-pixel confidence logit. We pass the logit
through a softplus to obtain the positive confidence $C_{v,t}(u) >
0$. Following DUSt3R~\cite{wang2024dust3r} and
MapAnything~\cite{keetha2025mapanything}, the loss is
confidence-weighted in log space, with Barron's robust
$\rho$~\cite{barron2019general}.

\mypar{Training Schedule}

We initialize from the SVD checkpoint used by
Ctrl-World~\cite{guo2025ctrl}, utilizing identical action
conditioning (per-frame end-effector pose). The combined training objective
is $\mathcal{L} = \mathcal{L}_{denoise} + \lambda_{pm}
\mathcal{L}_{pm}$. We train in two stages for stability. Stage 1
trains the joint RGB-depth model alone for 40,000 steps ($\lambda_{pm} = 0$),
allowing the model to learn the joint latent distribution. Stage 2
initializes the DPT head from VGGT and trains all components
jointly for an additional 50,000 steps ($\lambda_{pm} = 0.005$).

\section{Experiments}
\label{sec:eval}

We evaluate both contributions empirically: the calibration quality of
DROID-3D against a per-scene baseline, and the world model prediction
accuracy on held-out trajectories.

\mypar{DROID-3D evaluation setup and baseline.}
We evaluate calibration on 5 DROID labs ($\sim$1{,}000 scenes total).
Our baseline is a per-scene refinement procedure modeled
on PointWorld~\cite{huang2026pointworld}: VGGT~\cite{wang2025vggt}
extrinsic initialization followed by per-scene optimization against
robot depth, with S$^2$M$^2$~\cite{min2025s2m2} substituted as the
depth source to match ours. We use four geometric proxy metrics:
ext--ext (\textbf{EE}) and wrist--ext (\textbf{WE}) reprojection error
on MAGSAC~\cite{barath2020magsac++} gold inliers, robot depth error
(\textbf{RD}) between S$^2$M$^2$ and URDF-rendered depth, and
\textbf{Mask IoU} between the URDF-rendered robot mask and a
RoboEngine~\cite{yuan2025roboengine} segmentation. Full definitions are
in Appendices~\ref{app:calib_baseline}--\ref{app:calib_metrics}. 

Additionally, we replicated the DROID setup in our lab and recorded $5$ ground truth (GT) poses of the cameras as well as the robot joint offsets using a ChArUco board. Using this GT setup, we measure the average rotation error and translation error in the optimized poses. We test the pose accuracy using $50$ synthetic scenes from the REALM simulator~\cite{sedlacek2026realm}.

\mypar{DROID-3D calibration accuracy.}
Table~\ref{tab:calibration} reports results on the full evaluation set
and on the subset where the baseline converges. The gap is large on
every axis: 23$\times$ improvement on EE, 13$\times$ on WE, 8$\times$ on RD, and
+37 absolute points on IoU; 
This substantial gap remains even when restricting evaluation to the subset of scenes where the baseline successfully converges.
Beyond accuracy, the baseline fails to converge on $\sim$18\% of
scenes (180 of 995); our coupled optimization has no analogous per-scene
failure mode (per-factor ablations: Table~\ref{tab:ablation-pose}; qualitative: Figure~\ref{fig:qual_pose_all}). 

\mypar{Accuracy to GT poses.}
On our GT setup, from teleoperation episodes alone, our approach recovers extrinsics to \textbf{4.9\,mm/0.39$^\circ$ median error} and hand-eye to \textbf{3\,mm/0.3$^\circ$} across all five capture sets; joint offsets are recovered to 0.14$^\circ$ mean error across all seven joints. On the same set-up, the PointWorld-style per-scene baseline obtains 72\,mm / 3.0$^\circ$ median error, an order of magnitude worse than ours. In REALM simulation with exact GT poses (50 configurations, varied viewpoints), our approach recovers extrinsics with \textbf{6.4\,mm/0.3$^\circ$} median error.

  \definecolor{consistencyRow}{HTML}{E8F0F8}
  \definecolor{groundingRow}{HTML}{FBEFD9}
  \begin{table}[b]
  \vspace{-1em}
  \centering
  \small
  \setlength{\tabcolsep}{6pt}
  \renewcommand{\arraystretch}{1.15}
  \caption{\textbf{Calibration quality} on $\sim$1{,}000 evaluation
  scenes. Blue rows: 2D multi-view consistency (medians on
  joint-MAGSAC gold inliers). Amber rows: 3D grounding to the robot frame. The right block restricts to the subset where the baseline's
  per-scene refinement converges.}
  \label{tab:calibration}
  \begin{tabular}{l cc c cc}
  \toprule
   & \multicolumn{2}{c}{\textit{All scenes} ($n = 995$)} & & \multicolumn{2}{c}{\textit{PointWorld-converged subset} ($n = 815$)} \\
  \cmidrule(lr){2-3} \cmidrule(lr){5-6}
  \textbf{Metric} & \textbf{PointWorld~\cite{huang2026pointworld}} & \textbf{Ours} & & \textbf{PointWorld~\cite{huang2026pointworld}} & \textbf{Ours} \\
  \midrule
  \rowcolor{consistencyRow} EE median (px) $\downarrow$  & 5.84   & \textbf{0.25}  & & 5.67   & \textbf{0.25} \\
  \rowcolor{consistencyRow} WE median (px) $\downarrow$  & 13.19  & \textbf{1.03}  & & 12.52  & \textbf{1.02} \\
  \rowcolor{groundingRow}   RD median (mm) $\downarrow$  & 111.3  & \textbf{14.0}  & & 104.6  & \textbf{13.9} \\
  \rowcolor{groundingRow}   Mask IoU mean $\uparrow$     & 0.440  & \textbf{0.810} & & 0.458  & \textbf{0.819} \\
  \bottomrule
  \end{tabular}
  \end{table}

\begin{figure}[t]
    \centering
    \begin{subfigure}{0.49\linewidth}
        \includegraphics[width=\linewidth]{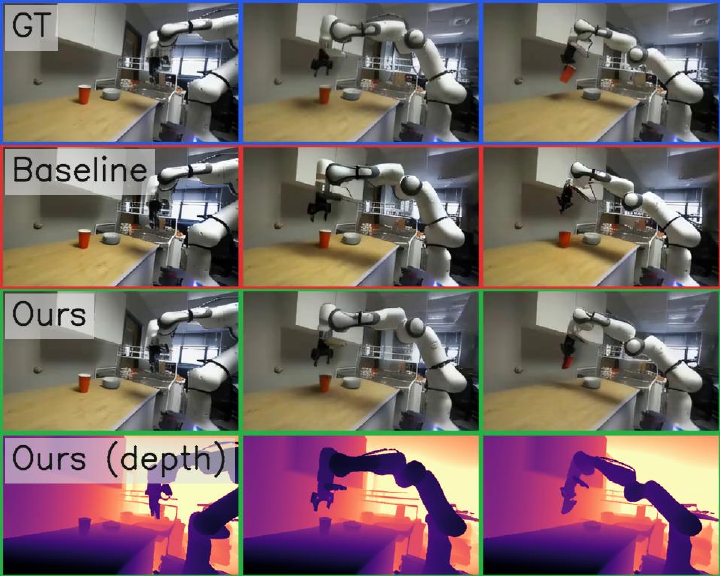}
    \end{subfigure} \hfill
    \begin{subfigure}{0.49\linewidth}
        \includegraphics[width=\linewidth]{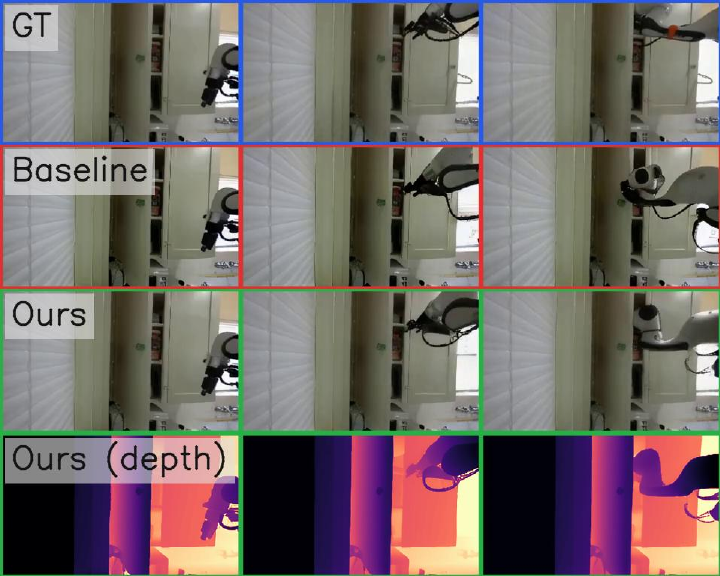}
    \end{subfigure}
    \caption{Qualitative Result: We show two rollouts by the baseline RGB model
and DepthWorld, which also produces high-quality depth.
\textbf{Left:} The RGB-only model fails to simulate the robot
grasping the red cup as in the GT trajectory. DepthWorld accurately
simulates the interaction and produces an aligned depth map.
\textbf{Right:} Without depth supervision, the RGB-only world model
cannot distinguish the open wardrobe door from a solid surface and
simulates the arm colliding with it. DepthWorld accurately generates
the arm going inside the wardrobe by accounting for the depth of the
arm and the door.
    {\bf See the project website for additional results, including videos.}
    }
    \label{fig:qual-res}
\end{figure}

\mypar{World model evaluation setup and metrics.}
We evaluate on 256 held-out trajectories from DROID with 10 consecutive
autoregressive rollouts. We compare three variants:
\textbf{RGB} is the Ctrl-World~\cite{guo2025ctrl}
backbone (multi-view RGB only); \textbf{RGB+D} is our joint
RGB-depth model; \textbf{RGB+D PM-DPT} adds the point-map auxiliary
head. We report PSNR, SSIM~\cite{ssim}, and LPIPS~\cite{lpips} for
RGB, and AbsRel, RMSE, and $\delta_1$ for depth, each computed
separately for external and wrist views. Depth metrics are against the depth obtained in Sec.~\ref{sec:dataset}.
Full protocol is in
Appendix~\ref{app:wm_eval}.

\mypar{Joint depth prediction improves RGB.}
Table~\ref{tab:wm_quality} reports both modalities on the held-out
set, with qualitative rollouts shown in Figure~\ref{fig:qual-res}.
Adding depth as a joint prediction target through spatial latent
tiling delivers a clear gain on RGB quality: RGBD models outperform the
RGB-only baseline on all three RGB metrics
with +1.48\,dB PSNR on external views and +1.02\,dB on wrist. 

\mypar{Point-map supervision and DROID-3D's downstream value.}
Adding the point-map auxiliary loss (PM-DPT) further refines depth accuracy—especially on the challenging wrist view -- while fully maintaining RGB generation quality. This suggests that mapping RGB and depth into a shared, spatially tiled latent space already encourages the U-Net to learn strong cross-modality constraints. Consequently, the robot-frame auxiliary loss acts primarily as a fine-grained 3D regularizer rather than the sole driver of geometric learning. More broadly, PM-DPT serves as a working instance of how DROID-3D's robot-grounded annotations can drive spatial supervision beyond independent per-view depth. This framework opens the door for future extensions, such as enforcing strict spatiotemporal (4D) consistency or enabling downstream policies to extract unified 3D representations directly from the world model's internal latents.

\mypar{Cross-view consistency improvements.} We measure two quantities: \emph{(i) Cross-view reprojection:} We warp one generated external view into the other using GT depth and poses. RGB+D model beats the RGB baseline model by \textbf{+1.0\,dB} PSNR (17.60 vs.\ 16.60), and our PM-DPT head always improves by $+0.05$--$0.08$\,dB over RGB+D. \emph{(ii) Cross-view generated-depth agreement:} 
Our PM-DPT model reduces external-view projected depth disagreement by $2.6$\,mm compared to our RGBD model.

\mypar{Comparison with other baselines.} 
We compare against the existing world models for manipulation that predict future 3D geometry rather than RGB alone: PointWorld (PW)~\cite{huang2026pointworld}, which predicts 3D scene flow and is the only action-conditioned 3D world model trained on DROID, and TesserAct~\cite{zhen2025tesseract}, which predicts RGB-D-normal video; iMoWM~\cite{zhang2025imowm} has no public code. TesserAct is text-conditioned and single-view, so it is incomparable as is; we adapted it to action conditioning and trained it on the same DROID split. At 50k steps it trails our model: external PSNR 19.43 vs 24.07 (ours), AbsRel 0.154 vs 0.074 (ours); see Appendix~\ref{app:tesseract_cmp} for the adaptation and protocol. For PW, whose output is a point cloud rather than depth images, we compare predicted 3D trajectories of scene points (robot excluded), each model scored on its own depth target; see Appendix~\ref{app:pointworld_cmp}, Table~\ref{tab:pointworld_cmp}, and Figure~\ref{fig:pw_qual}. Tracked-point $\ell_2$ (2$\to$8\,s): PW \textbf{23}$\to$56\,mm, ours 31$\to$\textbf{46}\,mm. Chamfer: whole scene PW 8$\to$18\,mm vs.\ ours \textbf{8$\to$9}\,mm; moving region PW \textbf{21}$\to$58\,mm vs.\ ours 25$\to$\textbf{32}\,mm. 

Appendix~\ref{app:depthworld} ablates the training depth source (Table~\ref{tab:depth_source}), the point-map supervision extrinsics (Table~\ref{tab:pm_extrinsics}), and the depth-integration architecture (Table~\ref{tab:tiling-vs-dualbranch}). 

Depth quality is strongly view-dependent across both depth-augmented
variants: $\delta_1 > 0.94$ on external views but $\approx 0.82$ on
wrist due to the wrist camera's rapid motion, proximity to the
manipulated object, and frequent self-occlusion by the arm.

\mypar{Limitations.}
As an autoregressive rollout model, DepthWorld inherits the exposure bias and quality degradation common to such models~\cite{guo2025ctrl}. However, recent works like PersistWorld~\cite{bardhan2026persistent} have shown that RL post-training can significantly improve rollout stability. The availability of high-quality 3D annotations in DROID-3D, combined with the native 3D-consistent outputs of DepthWorld, opens exciting new frontiers for geometric reward design in future world modelling.

\begin{table}[t]
  \centering
  \small
  \setlength{\tabcolsep}{6pt}
  \renewcommand{\arraystretch}{1.15}
  \caption{\textbf{Ablation of world model prediction quality on held-out
  trajectories.} Depth metrics are against obtained depth from Sec.~\ref{sec:dataset}. The RGB-only baseline does not predict depth
  (---). Joint depth prediction (RGB+D) improves all RGB metrics on
  both view classes; the point-map auxiliary loss (PM-DPT) further refines depth accuracy while maintaining RGB generation quality, specially on wrist metrics.}
  \begin{tabular}{ll cccccc}
  \toprule
   & & \multicolumn{3}{c}{RGB quality} & \multicolumn{3}{c}{Depth quality} \\
  \cmidrule(lr){3-5} \cmidrule(lr){6-8}
  \textbf{View} & \textbf{Model} & PSNR$\uparrow$ & SSIM$\uparrow$ & LPIPS$\downarrow$ & AbsRel$\downarrow$ & RMSE$\downarrow$ & $\delta_1{\uparrow}$ \\
  \midrule
  \multirow{3}{*}{External} & RGB          & 22.63          & 0.8170          & 0.0930          & ---             & ---            & ---            \\
                            & RGB+D        & \underline{24.09}          & \textbf{0.8333} & \textbf{0.0889} & \textbf{0.0765} & 0.220          & 0.9432         \\
                            & RGB+D PM-DPT & \textbf{24.11} & \textbf{0.8333} & \underline{0.0891}          & 0.0782          & \textbf{0.219} & \textbf{0.9434}\\
  \midrule
  \multirow{3}{*}{Wrist}    & RGB          & 16.98          & 0.5737          & 0.3228          & ---             & ---            & ---            \\
                            & RGB+D        & \underline{17.96}          & \underline{0.6106}          & \underline{0.3137}          & 0.2230          & 0.150          & 0.8182         \\
                            & RGB+D PM-DPT & \textbf{18.00} & \textbf{0.6129} & \textbf{0.3131} & \textbf{0.2182} & \textbf{0.148} & \textbf{0.8210}\\
  \bottomrule
  \end{tabular}
  \label{tab:wm_quality}
  \vspace{-3mm}
\end{table}

\section{Conclusion}
\label{sec:conclusion}
We introduced two contributions toward 3D-consistent world models for
robot manipulation. \textbf{DROID-3D} recalibrates the DROID dataset
into a 3D corpus with dense metric depth and robot-grounded
multi-view extrinsics across more than 70,000 episodes, through a
factor graph that pools each robot's episodes against shared
kinematic parameters. \textbf{DepthWorld} consumes this signal in a
Stable Video Diffusion-based world model that jointly predicts
multi-view RGB and depth via spatial latent tiling. The central empirical finding is that at equal training budget, joint RGB-depth prediction improves RGB by +1.48\,dB PSNR over an RGB-only baseline, while simultaneously yielding metric depth for downstream geometric reasoning. Together, DROID-3D and DepthWorld open a path toward world models whose rollouts compose into a single coherent 3D scene---the property all downstream uses, policy evaluation, improvement, and planning, depend on.

\acknowledgments{This work was supported by the European Union's Horizon Europe projects AGIMUS (No. 101070165), euROBIN (No. 101070596), ERC FRONTIER (No. 101097822), ELIAS (No. 101120237), ELLIOT (No. 101214398), ČVUT Starting grant "DREAM-ACT" (Project ID CVUT-StG-26-089), and CTU Future Fund (Project ID: CVUT-BrF-26-22825M). This work was also supported by the EU’s Horizon Europe Programme under the Grant agreement No. 10113667 (CLARA Project), and was co-funded by the EU from the Operational Programme Jan Amos Komenský (OP JAK) (project ``Center for Artificial Intelligence and Quantum Computing in System Brain Research'', reg. no. CZ.02.01.01/00/23\_029/0008437). Compute resources and infrastructure were supported by the Ministry of Education, Youth and Sports of the Czech Republic through the e-INFRA CZ (ID:90254).}



\bibliography{example}  
\clearpage
\addtocontents{toc}{\protect\setcounter{tocdepth}{2}}

 \newpage
 \appendix

\setcounter{page}{1}
\setcounter{figure}{0}
\setcounter{table}{0}
\setcounter{equation}{0}
\renewcommand{\thefigure}{A\arabic{figure}}
\renewcommand{\thetable}{A\arabic{table}}
\renewcommand{\theequation}{A\arabic{equation}}

\addtocontents{toc}{\protect\setcounter{tocdepth}{2}}    
\renewcommand{\contentsname}{Supplementary Material for DepthWorld: 3D World Model for Robot Manipulation}                   
\tableofcontents

\vspace{5mm}

This supplementary material is organized as follows.
Appendix~\ref{app:stage1} and Appendix~\ref{app:stage2} detail the two stages of
the DROID-3D calibration pipeline: per-view metric depth and multi-view
consistency (Stage~1), and the robot-grounded joint factor graph with its
explicit residual definitions, hyperparameters, and discarded design choices
(Stage~2). Appendix~\ref{app:depthworld} covers the DepthWorld architecture and
training: depth preprocessing, spatial latent tiling, the point-map head, the
training objective, the optimization schedule, and the inference rollout.
Appendix~\ref{app:exp} specifies the calibration baseline, the geometric
calibration metrics, the world-model evaluation protocol, and the protocols
for the comparisons with PointWorld and TesserAct.

\section{Stage 1: Initial Multi-View Calibration}
\label{app:stage1}

\subsection{Per-View Metric Depth}
\label{app:depth}

\begin{figure}[tp]
    \centering
    \includegraphics[width=0.9\linewidth]{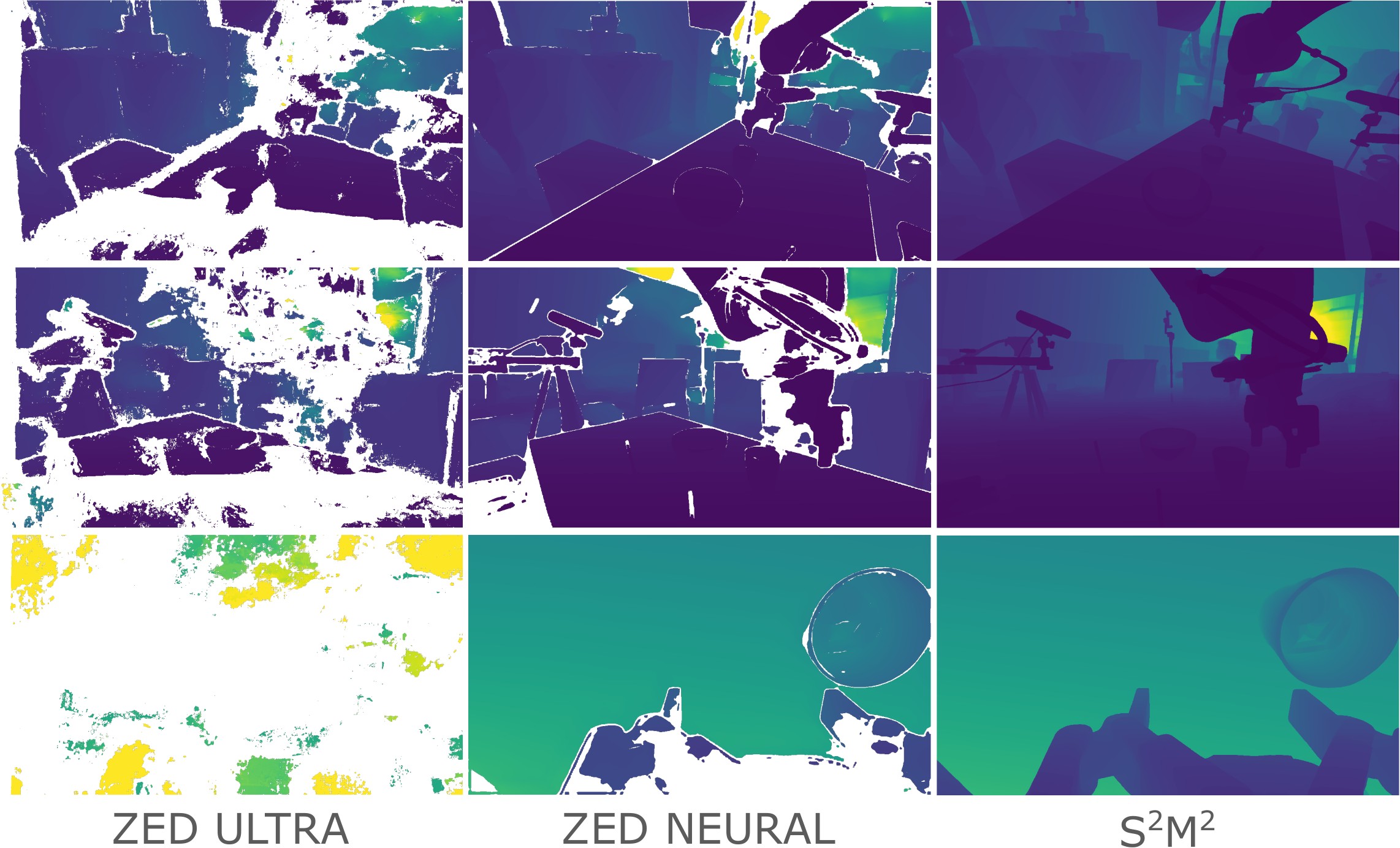}
    \caption{\textbf{ZED SDK vs S$\mathbf{^2}$M$\mathbf{^2}$}. The classical ZED ULTRA algorithm fails completely in the wrist cam view and produces sparse depth for the external views. ZED NEURAL produces comparatively denser results but fails at the boundaries and edges of objects. S$^2$M$^2$ produces denser and sharper results and outperforms both the alternatives.}
    \label{fig:zed-vs-s2m2}
\end{figure}

Each DROID camera (two table-mounted ZED~2 external cameras and a wrist-mounted
ZED~Mini) is a calibrated stereo pair with a known baseline, so metric depth
reduces to predicting per-pixel disparity. For a pixel $u$ with predicted
disparity $d(u)$ we obtain metric depth by direct triangulation,
\begin{equation}
  z(u) = \frac{f\,b}{d(u)},
  \label{eq:disp2depth}
\end{equation}
with focal length $f$ and baseline $b$ from the factory stereo calibration. The
choice of disparity estimator critically impacts the downstream photometric
reprojection losses, as sharper depth maps produce tighter geometric
correspondences. We use the global-matching stereo network
$S^2M^2$~\cite{min2025s2m2}, and evaluated three families of alternatives on
representative manipulation scenes:

\paragraph{Stereo SDKs (ZED \textsc{Ultra}, \textsc{Neural}).}
Classical semi-global matching (SGM) algorithms yield metric scale but fail on
the texture-poor tabletops and specular surfaces of the robot arm, resulting in
sparse depth maps with characteristic holes. The learned ZED \textsc{Neural}
mode is denser but remains semi-dense and is outperformed by $S^2M^2$ in both
coverage and boundary fidelity. See Figure~\ref{fig:zed-vs-s2m2} for qualitative examples
of the performance. Additionally, we found that on our GT setup with GT poses, S$^2$M$^2$ (ours) beats ZED NEURAL on robot-surface depth against the URDF render on all sets (median errors 11.2--15.3\,mm vs.\ 12.6--16.9\,mm).

\begin{figure}[tp]
    \centering
    \includegraphics[width=0.9\linewidth]{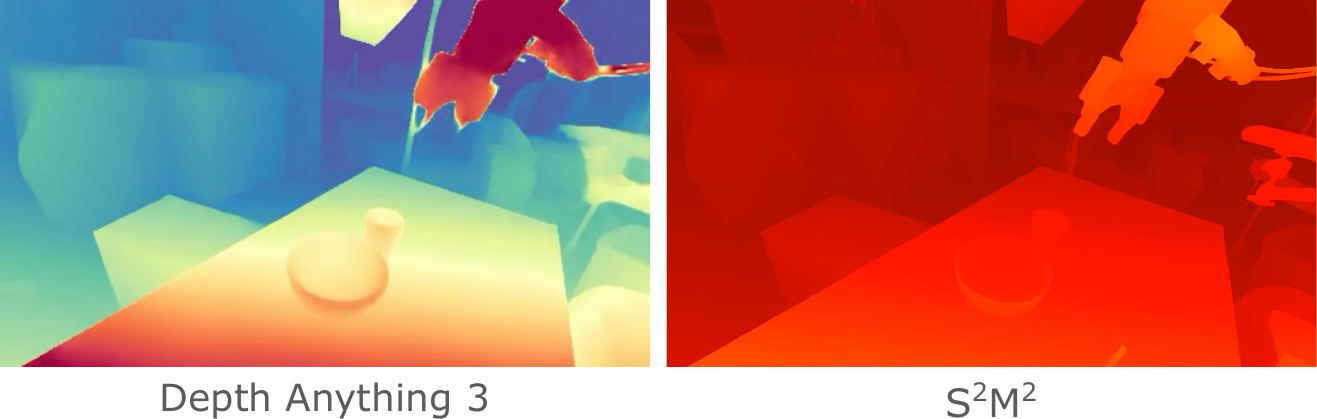}
    \caption{\textbf{Depth Anything 3 (DA3) vs S$\mathbf{^2}$M$\mathbf{^2}$}. DA3 produces generates noisy and overly smoothed depth maps, which fail to capture the details of the objects. The two methods are visualized with a different colormap (depth for DA3 vs disparity for S$^2$M$^2$).}
    \label{fig:da3-vs-s2m2}
\end{figure}

\paragraph{Feed-Forward Monocular Depth.}
Feed-forward monocular models, such as ~\cite{lin2025depth}, provide dense, visually
pleasing depth but suffer from an inherent affine scale ambiguity. This lack of
reliable metric scale causes severe misalignment during multi-view
triangulation. In our specific case with Depth Anything 3 (DA3), we find that
it generates noisy and overly smoothened depth maps, which fails to capture 
the details of the objects (see Figure~\ref{fig:da3-vs-s2m2}). We further note that DA3 is itself distilled from a stereo
teacher, and its own evaluation reports the stereo teacher to
be the stronger depth predictor---consistent with our preference for a stereo
source.

\begin{figure}[tp]
    \centering
    \includegraphics[width=0.9\linewidth]{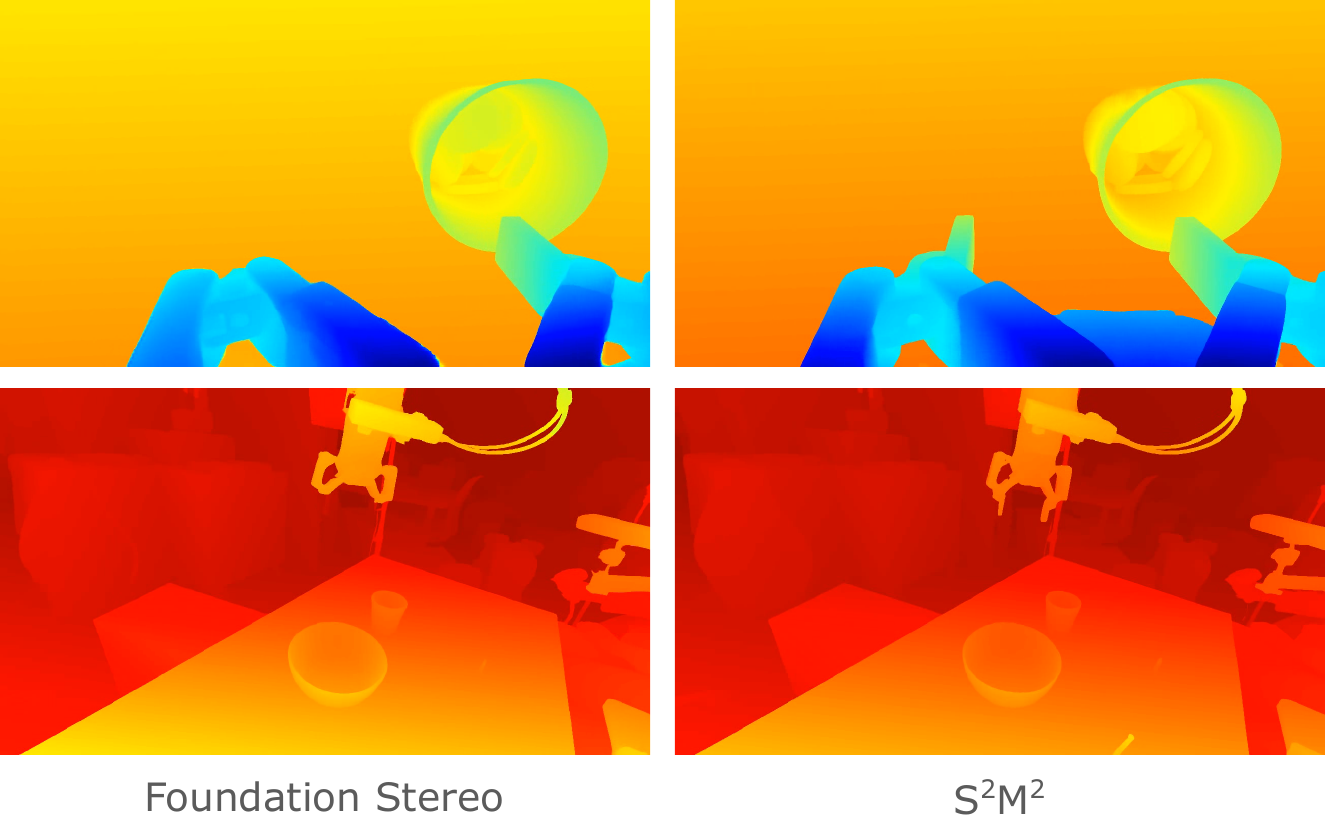}
    \caption{\textbf{Foundation Stereo vs S$\mathbf{^2}$M$\mathbf{^2}$}. We find that S$^2$M$^2$ produces more stable results. Notice the missing gripper on the left image and the higher quality details of the objects within the cup in the right image.}
    \label{fig:fs-vs-s2m2}
\end{figure}

\paragraph{Other Stereo Disparity methods.}
We also experimented with using FoundationStereo (FS)~\cite{wen2025foundationstereo} for recovering
the stereo disparity maps. However, we found that S$^2$M$^2$ more reliably produced
sharper results. Furthermore, S$^2$M$^2$ was significantly faster than FS at the $1280\times 720$p
high resolution, and did not require special Hierarchical models (as opposed to FS requiring Hiera~\cite{ryali2023hiera} models). 
For a qualitative comparison see  Figure~\ref{fig:fs-vs-s2m2}.

\subsection{Robust Multi-View Consistency Optimization}
\label{app:consistency}

\begin{figure}
    \centering
    \includegraphics[width=\linewidth]{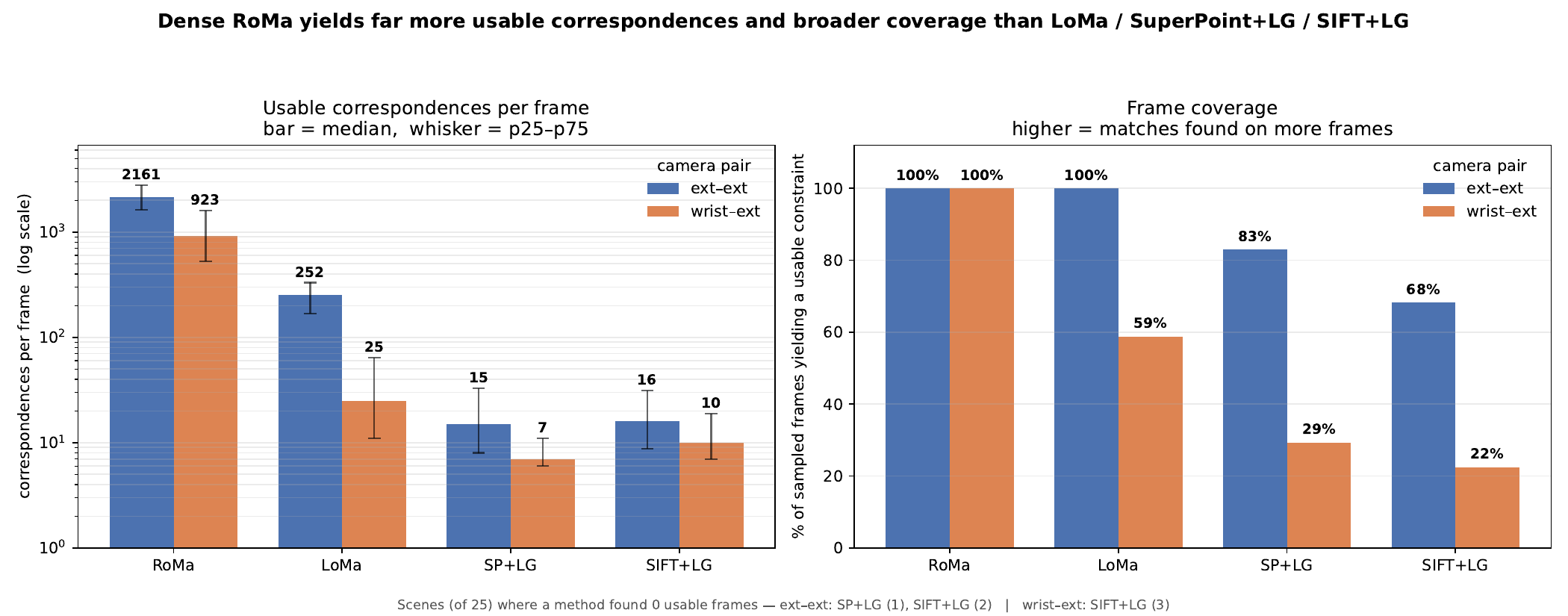}
    \caption{\textbf{RoMa-v2 vs others summary.} \textbf{Left:} We show the number of usable correspondences (inliers from MAGSAC) for each method. RoMa-v2 consistently produces an order of magnitude more correspondences. \textbf{Right:} We show the coverage of each method. RoMa-v2 has approximately $2\times$ the coverage on the extreme viewpoint pair of wrist and external cameras.}
    \label{fig:roma_summary}
\end{figure}

\paragraph{Correspondences and outlier rejection.}
Standard sparse keypoint detectors routinely fail to extract sufficient
correspondences in this domain due to the extreme viewpoint differences between
the fixed external cameras and the dynamic wrist camera, compounded by
textureless surfaces: SIFT~\cite{sift},
SuperPoint~\cite{superpoint}~+~LightGlue~\cite{lightglue}, and
LoMa~\cite{nordstrom2026loma} all return too few reliable matches for bundle
adjustment to converge, particularly on wrist--external pairs
(Figures~\ref{fig:feat_corr_1}--\ref{fig:feat_corr_3}). We instead rely on
RoMa~v2~\cite{edstedt2025roma} to extract dense, pixel-wise correspondences even
in textureless regions, robustly filtered using MAGSAC~\cite{barath2020magsac++}
fundamental-matrix estimation. Figure~\ref{fig:roma_summary} shows the count of correspondences and the coverage across the frames for a small set of $25$ samples.

\newcommand{\CorrPanels}[1]{%
  \begin{subfigure}[t]{0.49\linewidth}
    \centering
    \includegraphics[width=\linewidth]{images/7_appendix/correspondence_comparions/#1_RoMa.png}
    \caption{RoMa-v2~\cite{edstedt2025roma}}
  \end{subfigure}\hfill
  \begin{subfigure}[t]{0.49\linewidth}
    \centering
    \includegraphics[width=\linewidth]{images/7_appendix/correspondence_comparions/#1_LoMa.png}
    \caption{LoMa~\cite{nordstrom2026loma}}
  \end{subfigure}

  \vspace{0.6em}

  \begin{subfigure}[t]{0.49\linewidth}
    \centering
    \includegraphics[width=\linewidth]{images/7_appendix/correspondence_comparions/#1_SPLG.png}
    \caption{SuperPoint~\cite{superpoint} + LightGlue~\cite{lightglue}}
  \end{subfigure}\hfill
  \begin{subfigure}[t]{0.49\linewidth}
    \centering
    \includegraphics[width=\linewidth]{images/7_appendix/correspondence_comparions/#1_SIFTLG.png}
    \caption{SIFT~\cite{sift} + LightGlue~\cite{lightglue}}
  \end{subfigure}%
}

\begin{figure}[tp]
  \centering
  \CorrPanels{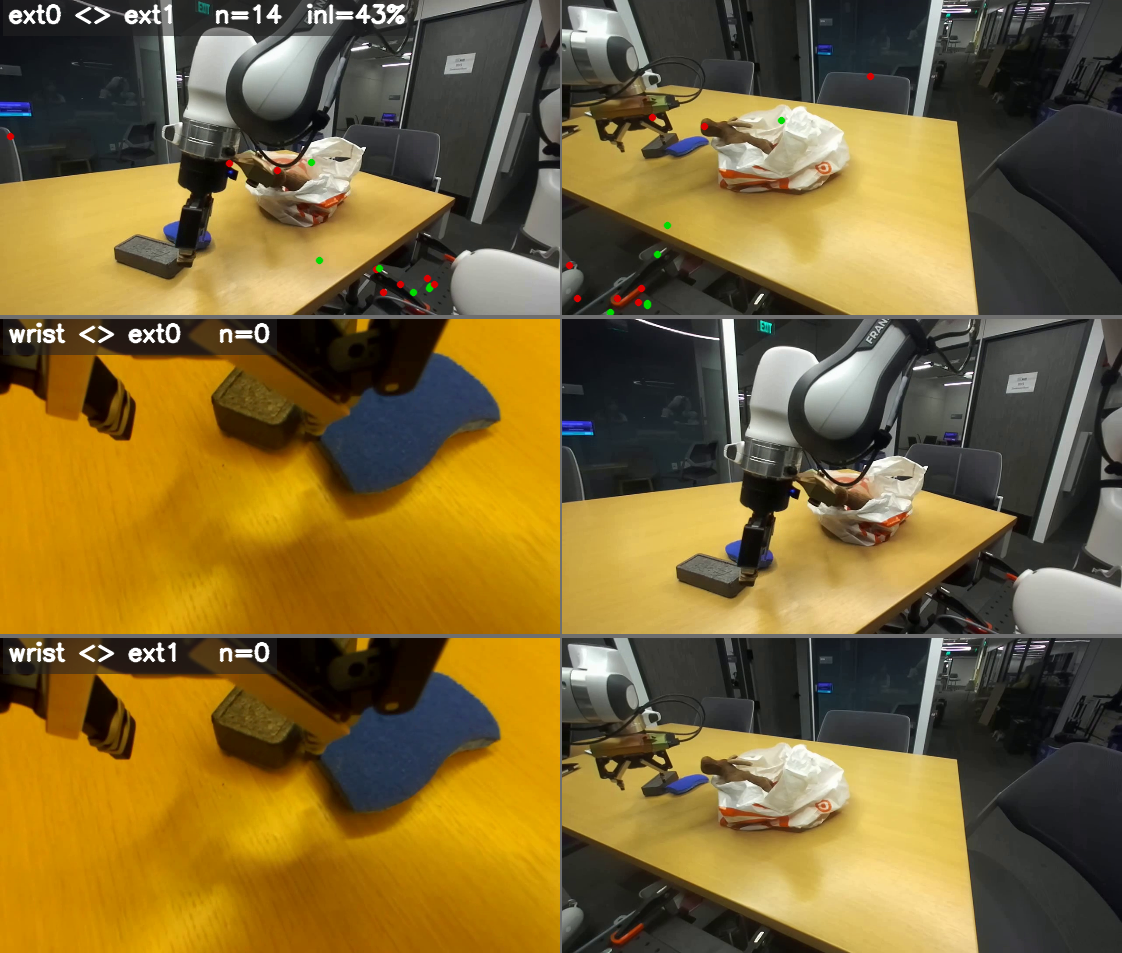}
  \caption{\textbf{Correspondence quality across feature matchers}. Each panel overlays the correspondences recovered by
  one matcher on a wrist--external camera pair; green points are
  MAGSAC~\cite{barath2020magsac++} fundamental-matrix inliers. The dense matcher
  RoMa-v2~(a) recovers accurate correspondences across the extreme viewpoint
  change, whereas the sparse matchers~(b--d) collapse to a handful of matches or
  fail outright on the textureless and specular regions that dominate the scene.}
  \label{fig:feat_corr_1}
\end{figure}

\begin{figure}[tp]
  \centering
  \CorrPanels{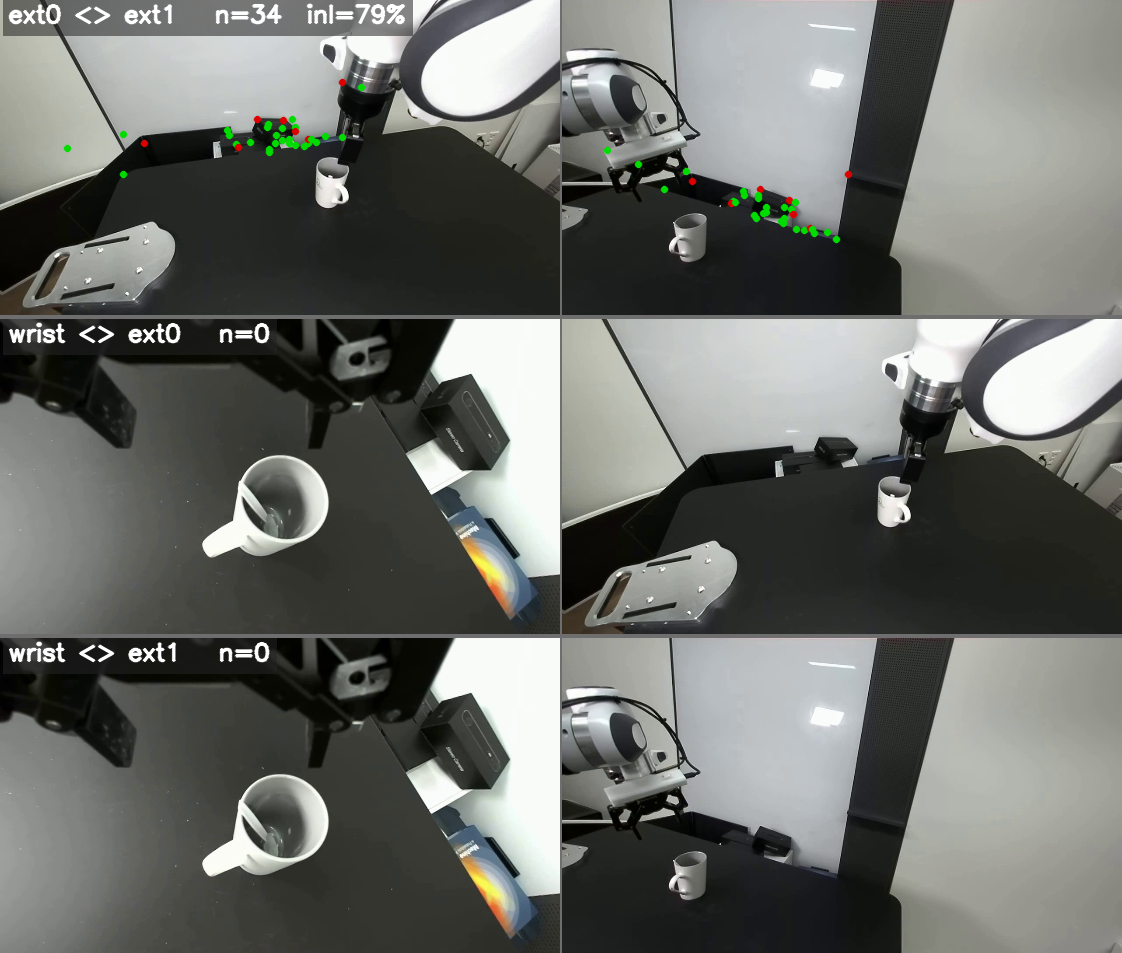}
  \caption{\textbf{Correspondence quality across feature matchers} (continued). Panels and color coding as in
  Fig.~\ref{fig:feat_corr_1}: (a)~RoMa-v2, (b)~LoMa, (c)~SuperPoint +
  LightGlue, (d)~SIFT + LightGlue; green points are MAGSAC inliers. RoMa-v2 again
  yields dense, accurate matches where the sparse matchers largely fail on the
  wrist--external pair.}
  \label{fig:feat_corr_2}
\end{figure}

\begin{figure}[tp]
  \centering
  \CorrPanels{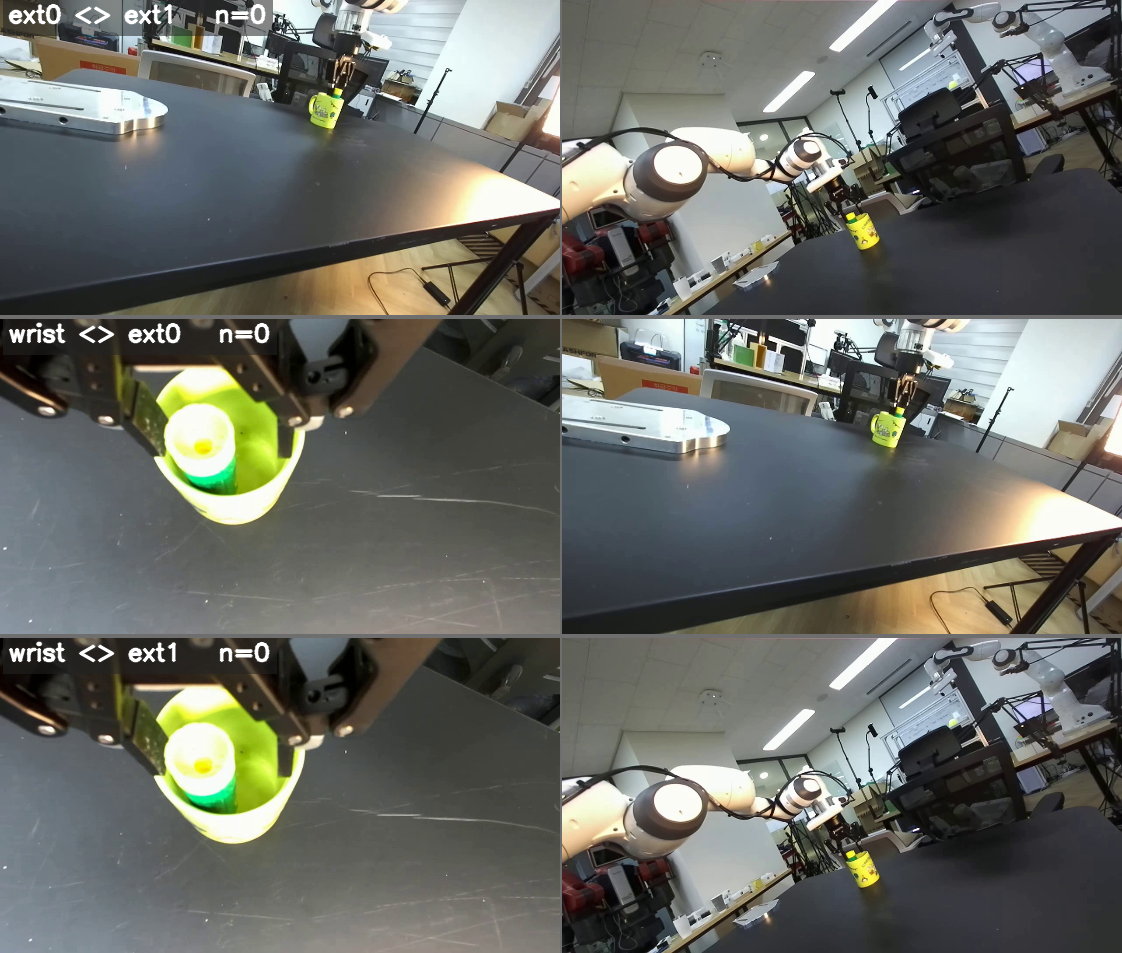}
  \caption{\textbf{Correspondence quality across feature matchers} (continued). Panels and color coding as in
  Fig.~\ref{fig:feat_corr_1}: (a)~RoMa-v2, (b)~LoMa, (c)~SuperPoint +
  LightGlue, (d)~SIFT + LightGlue; green points are MAGSAC inliers. The gap is
  consistent across scenes---RoMa-v2 sustains coverage in the wrist view while
  the other matchers degrade.}
  \label{fig:feat_corr_3}
\end{figure}

\paragraph{Two-pass bundle adjustment.}
To avoid suboptimal local minima caused by the noise disparity between
external--external and wrist--external view pairs, the bundle adjustment is
solved in two passes.
\emph{Pass~1} locks the two external cameras $(E_1, E_2)$ into a mutually
consistent pair: using the metric depth to lift matched pixels, we reproject
depth from one external view into the other and minimize the Cauchy photometric
residual. This reliably reaches sub-pixel triangulation error on the external
pair for the large majority of scenes. The resulting pair is internally
consistent but its global pose in the robot frame remains undetermined (a
$6$-DoF gauge freedom; metric scale is fixed by stereo depth).
\emph{Pass~2} fixes this external pair internally and resolves its global
$6$-DoF pose by anchoring it to the wrist camera's forward-kinematic trajectory.
The wrist pose at each frame is \emph{not} a free variable: it is taken from the
robot's kinematic chain---forward kinematics on the raw joint angles composed
with DROID's factory hand-eye prior---and held fixed in this pass. We solve for
the single global $SE(3)$ transform of the locked external pair that best
explains the wrist--external correspondences, using a looser Cauchy scale to
reflect the noisier wide-baseline matches. The output is a per-scene rig
approximately in the robot's base frame---approximate because it still inherits
the systematic hand-eye and joint-encoder errors of the factory priors, which
Stage~2 (Appendix~\ref{app:stage2}) removes.

\section{Stage 2: Robot-Grounded Joint Optimization}
\label{app:stage2}

The input to this stage is the set of visually consistent per-scene rigs from
Stage~1, together with the robot's URDF and raw joint encoder readings. The
output is a set of refined extrinsics anchored to the robot's base frame,
coupled with globally optimized robot-specific kinematic parameters (hand-eye
transform and joint encoder offsets).

\subsection{Kinematic Anchoring via Rendering}
\label{app:anchoring}

\begin{figure}
    \centering
    \includegraphics[width=0.8\linewidth]{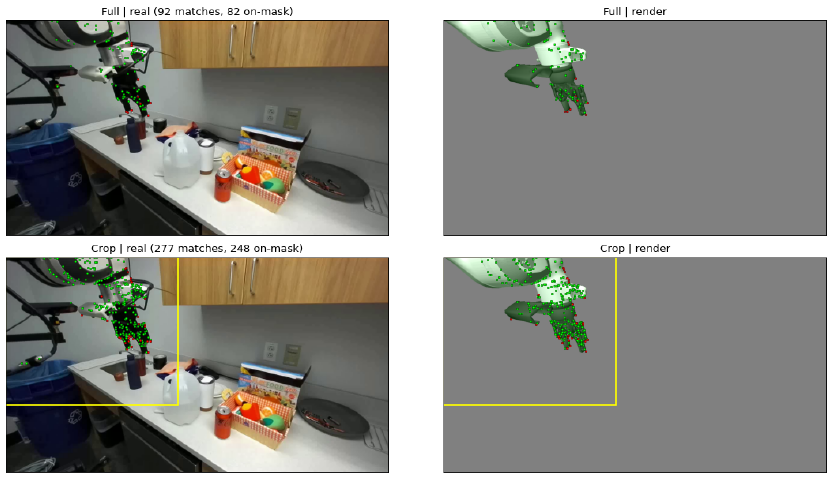}
    \caption{Effect of cropping the render and the real image before matching with LoMa-G. The yellow box indicates the crop region. The number of matches goes up from $92\rightarrow277$. Green dots indicate matches that are on the surface of the robot, red dots are for points that do not lie on the surface.}
    \label{fig:urdf-render}
\end{figure}

\begin{figure}
    \centering
    \includegraphics[width=0.8\linewidth]{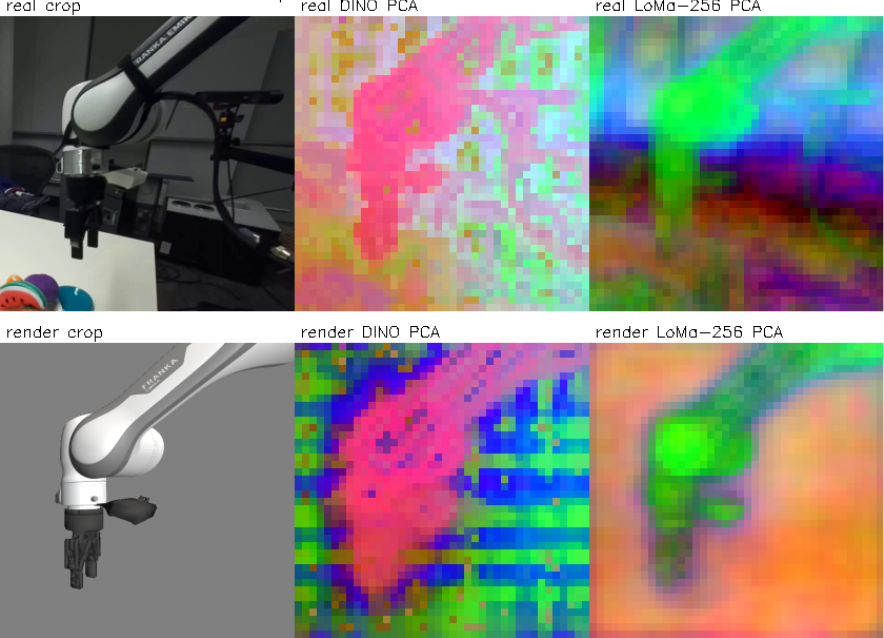}
    \caption{\textbf{Visualization of the PCA of the DINOv2 features and the LoMa features.} Here we show a comparison between the patch level features of the actual image, cropped around the robot arm, and a similar render of the robot URDF. The patch level features for both DINOv2 and LoMa match at the robot arm. However, the patch level features from LoMa contain more fine-grained details, since LoMa was specifically tuned for matching.}
    \label{fig:fm-render}
\end{figure}

To tie the rig to the physical robot, we render the arm into each external view
with \texttt{pyrender}, using the URDF, the current joint angles, and the
Stage~1 extrinsics. We extract $2$D--$3$D correspondences between the rendered
template and the real image with LoMa-G~\cite{nordstrom2026loma}: each match
links a rendered surface point---whose $3$D location in the robot frame is known
from the URDF and forward kinematics---to a real-image pixel, yielding a
robot-frame ground-truth signal independent of any cross-view photometric cue.
LoMa-G internally computes DINO~\cite{oquab2023dinov2} patch features; we store
these at the correspondences and PCA-reduce them to $64$ dimensions to support a
featuremetric alignment term~\cite{sarlin2021back,mikevstikova2026alignpose} (the $\text{f}$ term in
Eq.~\eqref{eq:objective}) that provides sub-pixel gradients in textureless
regions. Two practical details matter. First, we crop both the render and the
real image tightly around the rendered arm before matching: this raises both the
match count and the effective patch resolution ($92 \to 277$ matches in
Fig.~\ref{fig:urdf-render}). Second, we render with well-lit,
non-physically-based (non-PBR) shading, which produces more stable and dense
discrete correspondences for LoMa-G than physically-based rendering. Figure~\ref{fig:fm-render} shows the patch level features for rendered robot and the input image. We see that the patches between the render and the actual image align nicely. Note that the features produced by LoMa are higher resolution and align much better than raw DINOv2 features. Therefore, we use the LoMa generated features (which builds upon the DINOv2 features).

\subsection{Joint Factor Graph: Residual Definitions}
\label{app:residuals}

A single factor graph spans all $N_s$ episodes of one physical robot ($N_s$
ranges from ${\sim}1{,}000$ to ${\sim}9{,}000$). Each episode contributes
\emph{per-scene} $6$-DoF $SE(3)$ corrections $\delta T^s_{\text{ext1}},
\delta T^s_{\text{ext2}}$ to its two external cameras ($12$~DoF/episode), which
absorb scene-specific calibration noise. \emph{Shared} across all episodes are
the $6$-DoF hand-eye correction $\delta T_{\text{wrist}}$ and the $7$-DoF joint
encoder offsets $\delta\bm q$ ($13$~DoF total). The wrist camera carries no
per-episode degrees of freedom; at frame $t$ its pose is fully determined by the
kinematic chain of
Eq. (1).
For the largest robot groups this gives
$12 N_s + 13 \approx 1.08\times10^{5}$ parameters.

\paragraph{Notation.}
Let $K_c$ and $T^s_c \in SE(3)$ (camera-to-world) be the intrinsics and
extrinsics of camera $c$ in scene $s$, with $T^s_c = \delta T^s_c \,\hat T^s_c$
for the Stage-1 initialization $\hat T^s_c$. For pixel $\mathbf u$ (homogeneous
$\tilde{\mathbf u}$) with metric depth $D_c(\mathbf u)$, the back-projected
world point is $\mathbf X = T^s_c\,D_c(\mathbf u)\,K_c^{-1}\tilde{\mathbf u}$, and
$\pi(\cdot)$ denotes perspective projection,
$\pi([x,y,z]^\top)=[x/z,\,y/z]^\top$. All data terms enter the scaled-Cauchy
loss $\rho(\lVert r\rVert^2; c) = c^2\log\!\bigl(1+\lVert r\rVert^2/c^2\bigr)$ of
Eq. (2).

\paragraph{The five data terms.}
\begin{itemize}
  \item \textbf{Cross-camera, external--external} ($\text{ee}$). For a
  RoMa-v2 match $(\mathbf u,\mathbf u')$ between $E_1$ and $E_2$ in scene $s$,
  lifted via $S^2M^2$ depth $D_{E_1}(\mathbf u)$:
  \begin{equation}
    r^{\text{ee}}_{s,i} = \pi\!\Bigl(K_{E_2}\,(T^s_{E_2})^{-1}\,T^s_{E_1}\,
      D_{E_1}(\mathbf u)\,K_{E_1}^{-1}\tilde{\mathbf u}\Bigr) - \mathbf u'.
  \end{equation}

  \item \textbf{Wrist--external 2D reprojection} ($\text{2d}$). For a match
  $(\mathbf u_W,\mathbf u_E)$ between the wrist $W$ at frame $t$ and external
  $E_c$, with $T_W(t)$ given by the chain in Eq.~\eqref{eq:chain}:
  \begin{equation}
    r^{\text{2d}}_{s,i} = \pi\!\Bigl(K_{E_c}\,(T^s_{E_c})^{-1}\,T_W(t)\,
      D_W(\mathbf u_W)\,K_W^{-1}\tilde{\mathbf u}_W\Bigr) - \mathbf u_E.
  \end{equation}
  This term is Mahalanobis-whitened: the residual entering $\rho$ is
  $\Sigma_i^{-1/2}\,r^{\text{2d}}_{s,i}$, where $\Sigma_i$ is the per-match
  covariance derived from RoMa~v2's predicted certainty, so noisier
  wrist--external matches contribute proportionally less.

  \item \textbf{Wrist--external 3D metric lifting} ($\text{3d}$, written
  $r^{we\text{-}3d}$). 
  We use a metric $3$D residual
  lifting the same wrist--external match into the world frame on both sides:
  \begin{equation}
    r^{\text{3d}}_{s,i} =
      T_W(t)\,D_W(\mathbf u_W)\,K_W^{-1}\tilde{\mathbf u}_W
      - T^s_{E_c}\,D_{E_c}(\mathbf u_E)\,K_{E_c}^{-1}\tilde{\mathbf u}_E
      \quad \in \mathbb{R}^3\ (\text{m}).
  \end{equation}

  \item \textbf{URDF-rendered robot reprojection} ($\text{l}$). For a LoMa-G
  $2$D--$3$D match between a rendered robot-surface point
  $\mathbf X^{\text{rob}}_{s,i}(\delta\bm q)$ (in the world frame, depending on
  $\delta\bm q$ through forward kinematics) and a real-image pixel
  $\mathbf u_{s,i}$ in $E_c$:
  \begin{equation}
    r^{\text{l}}_{s,i} = \pi\!\Bigl(K_{E_c}\,(T^s_{E_c})^{-1}\,
      \mathbf X^{\text{rob}}_{s,i}(\delta\bm q)\Bigr) - \mathbf u_{s,i}.
  \end{equation}
  While $\mathbf X^{\text{rob}}$ moves with $\delta\bm q$, in practice we use $\delta\bm q = 0$ for rendering the mesh, since the correction is generally very minor.

  \item \textbf{Featuremetric alignment} ($\text{f}$). With $\phi_i\in\mathbb
  R^{64}$ the PCA-reduced DINO feature of rendered point
  $\mathbf X^{\text{rob}}_{s,i}$ and $F_{E_c}:\Omega\to\mathbb R^{64}$ the dense
  real-image feature field (Fig.~\ref{fig:fm-render}):
  \begin{equation}
    r^{\text{f}}_{s,i} = F_{E_c}\!\Bigl(\pi\bigl(K_{E_c}\,(T^s_{E_c})^{-1}\,
      \mathbf X^{\text{rob}}_{s,i}\bigr)\Bigr) - \phi_i .
  \end{equation}
\end{itemize}
All reprojection terms operate on the MAGSAC-filtered inlier sets described in
Appendix~\ref{app:consistency}.

\subsection{Ablations on the Pose Estimation Pipeline}
\label{app:pose_ablation}

  \definecolor{consistencyRow}{HTML}{E8F0F8}
  \definecolor{groundingRow}{HTML}{FBEFD9}
  \begin{table}[t]
  \centering \small
  \setlength{\tabcolsep}{6pt}\renewcommand{\arraystretch}{1.15}
  \caption{\textbf{Ablation relative to the full model} on $\sim$1{,}000 scenes.
  All rows are the full model with the listed modification. Blue: 2D consistency
  (gold-inlier median px). Amber: 3D grounding. \textbf{$+$AMP\,3D, $-$FM} is the
  deployed model.}
  \label{tab:ablation-pose}
  \begin{tabular}{l >{\columncolor{consistencyRow}}r >{\columncolor{consistencyRow}}r
                    >{\columncolor{groundingRow}}r  >{\columncolor{groundingRow}}r}
  \toprule
  \textbf{Full model\ $\pm$} & \textbf{EE\,$\downarrow$} & \textbf{WE\,$\downarrow$}
                             & \textbf{RD\,$\downarrow$} & \textbf{IoU\,$\uparrow$} \\
   & (px) & (px) & (mm) & (mean) \\
  \midrule
  \textit{Full (all factors)}               & 0.251 & 1.032 & 14.15 & 0.808 \\
  \midrule
  \multicolumn{5}{l}{\textit{Weight tweaks}}\\
  \quad $+$ amplify 3D                       & 0.251 & 1.027 & 14.02 & 0.808 \\
  \quad $+$ amplify FM                       & 0.250 & 1.038 & 14.09 & 0.804 \\
  \midrule
  \multicolumn{5}{l}{\textit{Leave-one-out (drop one factor)}}\\
  \quad $-$ featuremetric                    & 0.251 & 1.031 & 14.03 & 0.808 \\
  \quad $-$ wrist-3D                         & 0.251 & 1.032 & 14.05 & 0.808 \\
  \quad $-$ featuremetric \& 3D              & 0.251 & 1.032 & 14.07 & 0.809 \\
  \quad $-$ split $T^s_{ext,k}$                  & 0.320 & 1.082 & 14.12 & 0.804 \\
  \quad $-$ ext--ext factor                  & 0.662 & 0.982 & 14.58 & 0.804 \\
  \quad $-$ joint offset $\delta q$          & 0.251 & 1.214 & 14.68 & 0.806 \\
  \quad $-$ $\delta T_{wrist}$                  & 0.252 & 2.067 & 17.11 & 0.786 \\
  \quad $-$ $\delta T_{wrist}$ \& $\delta q$    & 0.252 & 2.217 & 17.10 & 0.786 \\
  \midrule
  \textbf{$+$ amplify 3D, $-$ FM\ (ours)}    & \textbf{0.251} & \textbf{1.027} & \textbf{14.01} & \textbf{0.809} \\
  \bottomrule
  \end{tabular}
  \end{table}

We ablate the factors of the joint optimization in Table~\ref{tab:ablation-pose}. From the table, it is clear that our joint factor graph-based optimization of the shared robot parameters (such as hand eye calibration and robot joint offsets) significantly improves the quality of the obtained poses.  

\subsection{Hyperparameters and Solver}
\label{app:hparams}

All data terms use the robust Cauchy loss with empirically-set scales
$c_{\text{2d}} = 3\sigma$ (whitened distance), $c_{\text{3d}} = 10$~mm, $c_{\text{l}} = 50$~px, and
$c_{\text{ee}} = 5$~px. The 3D
metric-lifting factor is up-weighted by $\lambda_{3d}=10^4$ to rebalance the unit
mismatch between its metre-scale residual ($O(10^{-2})$) and the pixel-scale
factors ($O(1\text{--}50)$);
the remaining $\lambda_k$ are
$1$. The Tikhonov regularization weight on the joint offsets is
$\lambda_q = 10^{-2}$. The deployed (``ours'') configuration amplifies the
$3$D term and omits the featuremetric term ($-$FM), corresponding to the
\mbox{``$+$amplify\,3D, $-$FM''} row of Table~\ref{tab:ablation-pose}. While the featuremetric term improved the robot depth by itself, we found that it was a bit redundant after amplifying the 3D loss term. However, we still include the term in our description since our framework is general and computes the feature metric loss, which could be applied to further refine after optimization. 

The graph is large but its Hessian is highly sparse: each episode's $12$-DoF
block couples only to the $13$ shared parameters, never to other episodes. We
exploit this with a custom GPU-accelerated Levenberg--Marquardt solver. Each
iteration applies the Schur complement to marginalize all per-scene blocks,
reducing the step to a dense $13\times13$ global solve over the shared kinematic
parameters, followed by parallel per-scene back-substitution for the extrinsics.
This drops the computational cost from cubic in $N_s$ to
$O(N_s\cdot 12^3 + 13^3)$, with a typical convergence wall time of
${\approx}12.5$~minutes per robot group; in practice data loading, not the
linear algebra, is the bottleneck.

\subsection{Qualitative results of DROID-3D Calibration Dataset vs Baseline}
\label{app:qual_pose_all}

\begin{figure}[pth]
    \centering
    \includegraphics[width=\linewidth]{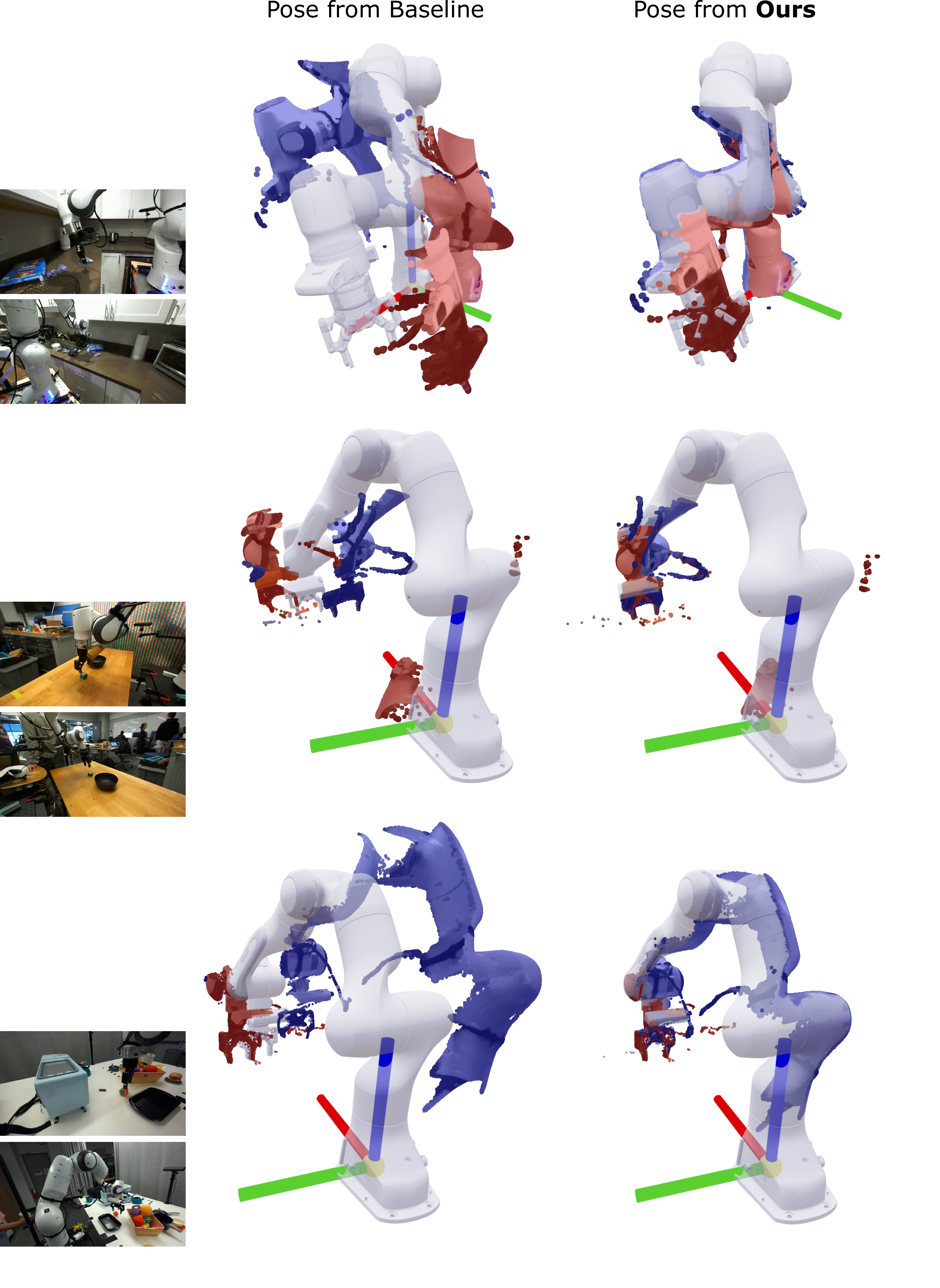}
    \caption{\textbf{Qualitative Results of the pose estimation} of PointWorld~\cite{huang2026pointworld} based baseline vs Ours. In all the cases our method produces significantly more aligned poses. The average robot depth error for PointWorld based approach on these examples is around 110-130mm while ours is around 15-20mm. We show only the points corresponding to segmented robot pixels in the image. In all the cases, it is clear that poses from the baseline method do not align, while they do for ours. Minor cleaning has been done to remove floaters due to imperfect segmentation.}
    \label{fig:qual_pose_all}
\end{figure}

We present some additional qualitative comparisons of the obtained poses by our method and the Point World based baseline method in Figure~\ref{fig:qual_pose_all}.

\section{DepthWorld Architecture and Training Details}
\label{app:depthworld}

\subsection{Depth Preprocessing Rationale (Supplement to Section 4.1)}
\label{app:depth_preproc}
Raw depth maps are converted to VAE-encodable grayscale images through a strict
four-step pipeline designed to keep the depth inside the distribution the Stable
Video Diffusion (SVD) VAE was pretrained on:

(1) We downsample to the $192 \times 320$ working resolution using unprojection
and z-buffer min-pooling rather than bilinear interpolation, which preserves
foreground geometry and prevents ``flying pixel'' artifacts at object
boundaries.

(2) Depths are clipped to the 95th percentile of the per-camera distribution
($\approx 3$ m for external, $\approx 2$ m for wrist) to bound the VAE input
while preserving manipulation-relevant content.

(3) A log transform compresses the dynamic range, allocating more
representational capacity to close-range objects where manipulation precision
matters most.

(4) Values are normalized to $[0, 1]$ to match the VAE's expected input range,
and the single channel is replicated to three channels.

\subsection{Spatial Latent Tiling vs. Other Depth Integration Architectures}
\label{app:tiling}
We opted for spatial latent tiling over dual-branch and channel expansion architectures. Dual-branch
methods require running a parallel U-Net for depth, linked via cross-connections
to the RGB branch, effectively doubling the parameter count and complicating the
training dynamics. 
Channel expansion, on the other hand, increase the number of channels in the input and output layers of the model to incorporate the added modality.
Our tiling approach leverages the observation that the SVD VAE
naturally encodes replicated single-channel depth, allowing us to place RGB and
depth side-by-side in a wider latent grid. This forces the single, unmodified
U-Net to jointly denoise both modalities, learning geometric consistency without major changes to the architecture.

We ablated this choice empirically. Table~\ref{tab:tiling-vs-dualbranch}
compares spatial latent tiling against a dual-branch model---a parallel U-Net for
depth with cross-connections to the RGB branch, in the style of
SyncVP~\cite{pallotta2025syncvp} trained for around $70$k
steps and evaluated under the autoregressive rollout protocol of
Appendix~\ref{app:wm_eval}. Different from~\cite{pallotta2025syncvp}, we do not use the same noise for both the modalities as our earlier experiments showed that it allowed the model to cheat in the EDM diffusion formulation. Furthermore, the model was first trained on RGB-only data for 50k steps and then continued for 20k steps with the dual branch. We would like to note that this is not a 1-1 comparison, but was done due to the additional compute requirements of the dual branch setup, which required more GPU hours to run through extensive FSDP parallelism. 
For the channel expansion architecture, we simply double the number of channels in the input and output layers of the model to incorporate the added modality. We train this in a single stage for 70k steps.
Spatial tiling performs better: it improves every RGB metric on both views (e.g.\ $+1.7$~dB PSNR on
external and $+1.5$~dB on wrist), and matches or improves depth quality, with the dual-branch only marginally ahead on
external RMSE. Therefore, this influenced our decision to avoid changing the architecture.
 
\begin{table}[t]
\centering \small
\setlength{\tabcolsep}{5pt}\renewcommand{\arraystretch}{1.15}
\caption{\textbf{Ablation: Spatial latent tiling vs.\ other integration architectures}, at $\sim70$k training steps
under the autoregressive rollout protocol (Appendix~\ref{app:wm_eval}). All are
joint RGB--depth models; depth metrics are against the VAE-round-tripped GT
(Appendix~\ref{app:depth_decode}). Spatial tiling matches or beats both the baseline integration architectures
architecture on nearly every metric and view despite leaving the U-Net unmodified.
Best per view/metric in \textbf{bold}, second best \underline{underlined}.}
\label{tab:tiling-vs-dualbranch}
\begin{tabular}{ll rrr rrr}
\toprule
 & & \multicolumn{3}{c}{RGB quality} & \multicolumn{3}{c}{Depth quality} \\
\cmidrule(lr){3-5}\cmidrule(lr){6-8}
View & Model & PSNR\,$\uparrow$ & SSIM\,$\uparrow$ & LPIPS\,$\downarrow$
              & AbsRel\,$\downarrow$ & RMSE\,$\downarrow$ & $\delta_1$\,$\uparrow$ \\
\midrule
External & Dual-branch            & 21.94 & \underline{0.8039} & \underline{0.1010} & 0.0861 & \underline{0.232} & \underline{0.9383} \\
         & Channel expansion  & \underline{22.22} & 0.8017 & 0.1021 & \textbf{0.0796} & \textbf{0.220} & \textbf{0.9414} \\
         & Spatial tiling (ours)  & \textbf{23.66} & \textbf{0.8252} & \textbf{0.0945} & \underline{0.0854} & 0.235 & \underline{0.9383} \\
\cmidrule(lr){1-8}
Wrist    & Dual-branch            & 16.05 & 0.5441 & 0.3554 & \underline{0.2523} & 0.161 & 0.7685 \\
         & Channel expansion  & \underline{16.34} & \underline{0.5593} & \underline{0.3422} & 0.2801 & \textbf{0.158} & \underline{0.7930} \\
         & Spatial tiling (ours)  & \textbf{17.58} & \textbf{0.5954} & \textbf{0.3276} & \textbf{0.2420} & \underline{0.159} & \textbf{0.8001} \\
\bottomrule
\end{tabular}
\end{table}

\subsection{Effect of the Depth Source on DepthWorld}
\label{app:depth_source}
DepthWorld is trained on S$^2$M$^2$ depth (Appendix~\ref{app:depth}). To quantify
how much the choice of depth source matters, we train two further models with
identical architecture and training protocol that instead use the ZED \textsc{Ultra}
and ZED \textsc{Neural} depth of the wrist and external cameras. All three models are
trained for $50$k steps and evaluated under the autoregressive rollout protocol of
Appendix~\ref{app:wm_eval}; Table~\ref{tab:depth_source} reports the results. Two
caveats apply when reading the table. The RGB metrics are directly comparable across
rows, since every model predicts the same RGB frames. The depth metrics are not: each
model is scored against the depth source it was trained on, so they measure how
faithfully a model reproduces its own supervision rather than its absolute metric
accuracy.

The depth source has a large effect. The model trained on ZED \textsc{Ultra} depth is
worst on every metric of both views: AbsRel rises from $0.074$ to $0.194$ on the
external view and from $0.266$ to $0.461$ on the wrist view, where \textsc{Ultra}
depth is particularly sparse and noisy (Figure~\ref{fig:zed-vs-s2m2}). Notably, the
poor depth target also degrades RGB prediction, by $2.0$~dB PSNR on the external view
and $1.4$~dB on the wrist view, even though the RGB supervision is unchanged. Sparse
and noisy depth is therefore not a neutral auxiliary signal but actively harms the
joint RGB--depth model, and a high-quality depth source is crucial for training it.

S$^2$M$^2$ and ZED \textsc{Neural} depth, in contrast, yield very similar models. The
RGB metrics agree to within $0.2$~dB PSNR on both views, and the depth metrics are
close, with S$^2$M$^2$ slightly ahead on the external view and on wrist $\delta_1$,
and ZED \textsc{Neural} slightly ahead on wrist AbsRel. This is expected: the two
depth sources are highly correlated (${\sim}0.97$ correlation between their depth
values), so they provide nearly the same training signal; where they differ is in
coverage and boundary sharpness (Appendix~\ref{app:depth}).

\begin{table}[t]
\centering \small
\setlength{\tabcolsep}{4pt}\renewcommand{\arraystretch}{1.15}
\caption{\textbf{Ablation: Depth source used to train DepthWorld}, at $50$k training
steps under the autoregressive rollout protocol (Appendix~\ref{app:wm_eval}). All are
joint RGB--depth models with identical architecture and training protocol; only the
depth maps used for training differ. Best per view/metric in \textbf{bold}, second
best \underline{underlined}.}
\label{tab:depth_source}
\begin{tabular}{@{}l cccc cccc@{}}
\toprule
 & \multicolumn{4}{c}{External} & \multicolumn{4}{c}{Wrist} \\
\cmidrule(lr){2-5}\cmidrule(lr){6-9}
$K{=}10$ rollout & PSNR$\uparrow$ & LPIPS$\downarrow$ & AbsRel$\downarrow$ & $\delta_1\uparrow$ & PSNR$\uparrow$ & LPIPS$\downarrow$ & AbsRel$\downarrow$ & $\delta_1\uparrow$ \\
\midrule
RGBD, S$^2$M$^2$ depth (ours) & \underline{24.07} & \textbf{0.0877} & \textbf{0.074} & \textbf{0.946} & \underline{17.68} & \underline{0.3270} & \underline{0.266} & \textbf{0.797} \\
RGBD, ZED NEURAL depth        & \textbf{24.26} & \underline{0.0881} & \underline{0.079} & \underline{0.939} & \textbf{17.77} & \textbf{0.3260} & \textbf{0.254} & \underline{0.794} \\
RGBD, ZED ULTRA depth         & 22.11 & 0.0947 & 0.194 & 0.830 & 16.28 & 0.3395 & 0.461 & 0.655 \\
\bottomrule
\end{tabular}
\end{table}

\subsection{Auxiliary Point-Map Head (Supplement to Section 4.2)}
\label{app:dpt_head}
The only auxiliary module is a single DPT~\cite{ranftl2021vision} point-map decoder.
It operates on the model's predicted clean latent for the future frames,
$\hat x_{0,f}$, not on intermediate U-Net activations. For each future frame and
view, the horizontal tiling is split back into its RGB and depth halves, and the
two $4$-channel latents are concatenated per view into an $8$-channel input of
shape $(B{\cdot}V,\, T_{\text{future}},\, 8,\, 24,\, 40)$.

A learned adapter then maps this latent into the DPT feature space, building a
four-scale feature pyramid from the $24\times40$ latent:
$24\times40 \rightarrow 12\times20 \rightarrow 6\times10 \rightarrow 3\times5$.
We reuse VGGT's~\cite{wang2025vggt} DPT fusion trunk and final regressor: the four
RefineNet fusion blocks (\texttt{refinenet1--4}) and the two output convolutions
(\texttt{output\_conv1,\,2}) of VGGT's point head are transferred and produce the
per-pixel $(X,Y,Z)$ point map in the robot base frame and a raw confidence
channel. Everything that adapts the Ctrl-World latents into that feature space is
trained from scratch: the input-clamp/latent input path, the stem, the three
pyramid stages, and the scale-projection convolutions (\texttt{layer1--4\_rn})
wherever VGGT's shapes do not match. The decoder upsamples to full image
resolution, outputting four channels at $192\times320$ (predicted tensor
$(B,\,V,\,T_{\text{future}},\,4,\,192,\,320)$). The raw confidence is mapped to a
strictly positive confidence via $C_{v,t}(\mathbf u) = 1 + \exp(\cdot)$,
following DUSt3R~\cite{wang2024dust3r}/MASt3R~\cite{leroy2024grounding}. 

Table~\ref{tab:pm_extrinsics} isolates the contribution of the recalibrated
extrinsics to this head: training PM-DPT with point-map targets built from DROID's
factory extrinsics instead of the refined DROID-3D extrinsics leaves RGB quality
unchanged but gives worse wrist-view depth (AbsRel $0.233$ vs.\ $0.220$).

\begin{table}[t]
\centering \small
\setlength{\tabcolsep}{4pt}\renewcommand{\arraystretch}{1.15}
\caption{\textbf{Ablation: Extrinsics used to build the point-map targets}, under the
autoregressive rollout protocol (Appendix~\ref{app:wm_eval}). Both are RGB+D PM-DPT
models trained identically; the robot-frame point-map targets are backprojected either
with DROID's factory extrinsics or with the refined DROID-3D extrinsics of
Appendix~\ref{app:stage2}. Best per view/metric in \textbf{bold}.}
\label{tab:pm_extrinsics}
\begin{tabular}{@{}l cccc cccc@{}}
\toprule
 & \multicolumn{4}{c}{External} & \multicolumn{4}{c}{Wrist} \\
\cmidrule(lr){2-5}\cmidrule(lr){6-9}
$K{=}10$ rollout & PSNR$\uparrow$ & LPIPS$\downarrow$ & AbsRel$\downarrow$ & $\delta_1\uparrow$ & PSNR$\uparrow$ & LPIPS$\downarrow$ & AbsRel$\downarrow$ & $\delta_1\uparrow$ \\
\midrule
PM-DPT, factory extrinsics & 24.10 & 0.0891 & \textbf{0.075} & \textbf{0.945} & 18.01 & 0.3131 & 0.233 & 0.817 \\
PM-DPT, refined extrinsics & \textbf{24.13} & \textbf{0.0890} & \textbf{0.075} & \textbf{0.945} & \textbf{18.03} & \textbf{0.3124} & \textbf{0.220} & \textbf{0.819} \\
\bottomrule
\end{tabular}
\end{table}

\subsection{Training Objective}
\label{app:objective_wm}
The combined objective is
\begin{equation}
  \mathcal{L} = \mathcal{L}_{\text{denoise}} + \lambda_{pm}\,\mathcal{L}_{pm},
  \qquad \lambda_{pm} = 0.005,
  \label{eq:wm_objective}
\end{equation}
where $\mathcal{L}_{\text{denoise}}$ is the standard Stable Video Diffusion
denoising objective applied to the joint tiled latent, with SVD's noise schedule
and loss weighting inherited unchanged.

The point-map term $\mathcal{L}_{pm}$ supervises the head's predicted point maps
$P_{v,t}$ against the ground-truth point maps $P^*_{v,t}$ (the world-frame /
robot-base XYZ point maps obtained by backprojecting the GT depth through the
camera intrinsics and the calibrated extrinsics of Appendix~\ref{app:stage2}),
with a per-pixel validity mask $m_{v,t}(\mathbf u)\in\{0,1\}$. Write
$\mathbf p = P_{v,t}(\mathbf u)$ and $\mathbf g = P^*_{v,t}(\mathbf u)$ for the
predicted and target $3$D points at a pixel. We compress the large dynamic range
of metric coordinates with a direction-preserving log map:
\begin{equation}
  \phi(\mathbf x) = \frac{\mathbf x}{\lVert\mathbf x\rVert}\,
    \log\!\bigl(1 + \lVert\mathbf x\rVert\bigr),
\end{equation}
and measure the residual in this space:
\begin{equation}
  \mathbf d = \phi(\mathbf p) - \phi(\mathbf g),
  \qquad
  s = \frac{\lVert \mathbf d\rVert^2}{c^2}, \quad c = 0.05 .
\end{equation}
The scaled residual $s$ is passed through Barron's general robust
loss~\cite{barron2019general} with shape $\alpha = 0.5$,
\begin{equation}
  \rho_\alpha(s) = \frac{|\alpha-2|}{\alpha}
    \left[\left(\frac{s}{|\alpha-2|} + 1\right)^{\!\alpha/2} - 1\right],
\end{equation}
and combined with the confidence $C = C_{v,t}(\mathbf u) = 1 + \exp(\cdot)$ in the
DUSt3R confidence-weighted form
\begin{equation}
  k_{v,t}(\mathbf u) = C\,\rho_\alpha(s) - \beta \log C,
  \qquad \beta = 0.2 .
\end{equation}
Two further operations are applied before aggregation. First, the largest $5\%$
of per-pixel terms $k$ within each (batch element, view) are dropped as outliers.
Second, each future-frame sample is weighted by its EDM loss weight, clamped by a
min-SNR-$\gamma$ of $5$. The
point-map loss is the masked, weighted mean over valid pixels,
\begin{equation}
  \mathcal{L}_{pm} =
    \frac{\sum m_{v,t}(\mathbf u)\,\sigma_w\,k_{v,t}(\mathbf u)}
         {\sum m_{v,t}(\mathbf u)},
\end{equation}
with the sum over the predicted future frames only.

\subsection{Training Schedule and Optimization}
\label{app:schedule}
We initialize from the Stable Video Diffusion checkpoint used by
Ctrl-World~\cite{guo2025ctrl} and inherit its action conditioning unchanged: a
per-frame $7$-dimensional end-effector action ($6$-DoF pose plus $1$ gripper
state). Training proceeds in two stages for stability, for $90{,}000$ steps in
total. Stage~1 trains the joint RGB--depth model alone for $40{,}000$ steps with
$\lambda_{pm}=0$, letting the model learn the joint tiled-latent distribution.
Stage~2 attaches the DPT head (VGGT fusion trunk and regressor reused, latent
adapter learned from scratch; Appendix~\ref{app:dpt_head}) and trains all
components jointly for a further $50{,}000$ steps with $\lambda_{pm}=0.005$.
We optimize with AdamW at a constant learning rate of $10^{-5}$ (no schedule) and
a global batch size of $64$. We maintain an exponential moving average (EMA) of
the weights with decay $0.9999$, and all reported results use the EMA model.
Total training takes approximately two days on two H200 nodes.

\subsection{Inference and Autoregressive Rollout}
\label{app:inference}
At inference the model denoises the joint tiled latent over a temporal window of
$11$ frames ($6$ history, $5$ future) at $5$~Hz, using $50$ denoising steps and a
classifier-free guidance scale of $2$ (matching Ctrl-World~\cite{guo2025ctrl}). The
history frames are sampled sparsely over the past, following
Ctrl-World~\cite{guo2025ctrl}, rather than as the six immediately preceding frames.
Each rollout starts from the first frame, with the history buffer initialized by
repeating it; the model then generates $10$ consecutive future chunks
autoregressively, feeding each predicted chunk back as history for the next
window. Reported metrics therefore reflect compounding prediction error from a
single-frame start, rather than teacher-forced single-step error.

\subsection{Depth Reference and De-normalization for Metric Evaluation}
\label{app:depth_decode}
To recover metric depth from a predicted depth tile we invert the four-step
preprocessing of Appendix~\ref{app:depth_preproc}: the depth tile is VAE-decoded,
the three replicated channels are averaged back to one, values are mapped from
$[0,1]$ back through the inverse of the normalization and the inverse log
transform, and the per-camera percentile clip is undone. The resulting depth is
metric. The reference is the same $S^2M^2$ ground-truth depth of
Appendix~\ref{app:stage2}, but passed through the VAE encode--decode round-trip
before scoring. Measuring against the VAE reconstruction of the GT depth, rather
than the raw GT, factors out the irreducible round-trip error that bounds any
latent-space predictor, so the depth metrics reflect prediction quality relative
to the best depth the latent representation can represent.

\section{Experimental Setup and Metrics}
\label{app:exp}

\subsection{Calibration Baseline Modifications (Supplement to Section 5.1)}
\label{app:calib_baseline}
To ensure a fair evaluation of our joint factor graph optimization against the
per-scene refinement baseline (modeled on PointWorld), we modified the
baseline's depth source. Instead of using FoundationStereo as in the original
PointWorld pipeline, we substituted our global-matching $S^2M^2$ depth. This
isolates the performance delta specifically to the structural difference in the
optimization (per-scene vs. robot-coupled joint graph) rather than a difference
in depth predictors.

\subsection{Geometric Calibration Metrics}
\label{app:calib_metrics}
Because ground-truth extrinsics are unavailable for DROID, we define four
geometric proxy metrics to evaluate calibration quality:
\begin{enumerate}
  \item \textbf{Ext-Ext (EE) Reprojection Error.} The median pixel reprojection
  error between the two external cameras, computed on a gold-inlier set
  extracted via MAGSAC enforcing epipolar and Kabsch-based 3D alignment.
  \item \textbf{Wrist-Ext (WE) Reprojection Error.} The analogous median pixel
  reprojection error measured between the dynamic wrist camera and the fixed
  external cameras.
  \item \textbf{Robot Depth Error (RD).} The median absolute difference (in
  millimeters) between the predicted $S^2M^2$ depth map and the depth obtained by
  rendering the robot URDF using the optimized extrinsics and joint angles.
  \item \textbf{Mask IoU.} The Intersection-over-Union between the URDF-rendered
  robot silhouette mask and a ground-truth robot segmentation mask produced by
  RoboEngine~\cite{yuan2025roboengine} on the real image.
\end{enumerate}

\subsection{World-Model Evaluation Protocol}
\label{app:wm_eval}

\paragraph{Training set and split.}
DepthWorld is trained on the calibrated episodes that constitute DROID-3D.
Starting from the publicly released raw DROID split (${\sim}75$k trajectories
with high-quality RGB, raw ZED SVO stereo, and factory camera parameters),
per-robot filtering for broken metadata or unrecoverable calibration leaves the
$71{,}100$ episodes across $28$ robots and $13$ labs reported in
Appendix~\ref{app:stage2}. DepthWorld trains on exactly these $71{,}100$
episodes. We hold out $1\%$ as a validation set and report world-model metrics on
$256$ trajectories sampled at random from it. Because the holdout is random at
the episode level, the held-out trajectories come from the same robots and labs
seen in training; these numbers therefore measure held-out-trajectory prediction
quality, not cross-embodiment or cross-lab generalization.

\paragraph{Baseline.}
The RGB baseline (\textbf{RGB} in Tables~1--2) is the Ctrl-World~\cite{guo2025ctrl}
backbone (multi-view RGB only), which we retrain rather than use off the shelf.
The released Ctrl-World checkpoint is trained on the $95$k DROID split stored in
RLDS format, whereas DROID-3D is constructed over the ${\sim}75$k publicly
released raw split. Retraining the RGB model on the same $71{,}100$-episode set,
split, architecture, and training budget as DepthWorld makes \textbf{RGB} and
\textbf{RGB+D} a controlled comparison in which the only difference is the depth
signal.

\paragraph{Metrics.}
For RGB we report PSNR, SSIM~\cite{ssim}, and LPIPS~\cite{lpips}.
For depth we report AbsRel, RMSE, and $\delta_1$, where
$\text{AbsRel} = \frac{1}{N}\sum |d - d^*|/d^*$, RMSE is in meters, and $\delta_1$
is the fraction of valid pixels with $\max(d/d^*,\, d^*/d) < 1.25$. Depth metrics
are evaluated in metric units without any per-frame scale-and-shift alignment,
against the VAE decoded GT depth reference of
Appendix~\ref{app:depth_decode}, and are masked to pixels with a valid depth
reference. All metrics are reported separately for external and wrist views, and
are averaged over all rollout frames.

\subsection{Comparison with PointWorld (Supplement to Section~\ref{sec:eval})}
\label{app:pointworld_cmp}

PointWorld~\cite{huang2026pointworld} and DepthWorld predict different things.
PointWorld is a 3D scene-flow model: given a set of seed points in the robot frame at
the anchor frame, it rolls out the future 3D position of every seed point. DepthWorld
predicts future depth images per camera, from which 3D points are obtained by sampling
the predicted depth at a pixel and unprojecting it with the calibrated intrinsics and
extrinsics. Neither output can be scored directly against the other, so the protocol
below lifts both to 3D points in the robot frame at the same tracked scene locations
and scores them there. Two consequences of this design should be kept in mind. First,
DepthWorld is only ever asked for the depth of a known pixel, which is an easier
question than PointWorld's task of predicting where an entire point set goes. Second,
DepthWorld predicts depth in the latent space of the SVD VAE, whose encode--decode
round trip alone has an error floor that PointWorld's point representation does not
have; how we account for this is described below.

\paragraph{Shared validation split.}
We use exactly the $256$ held-out windows of Table~\ref{tab:wm_quality}, recovered
from the same seeded random permutation of the validation set, so both models are
evaluated on identical clips. Each window consists of an anchor frame followed by $40$
future frames at $5$~Hz, i.e.\ an $8$~s horizon. Of these windows, $141$
contain moving scene points that survive the persistent point-set criterion described
below and are used for the moving-point metrics, and $147$ are used for the whole-scene
metric. We report a short horizon of $2$~s (frame $10$) and the full horizon of $8$~s
(frame $40$).

\paragraph{Tracker-based pseudo ground truth.}
DROID has no ground-truth 3D trajectories, so we build them from point tracks,
following the procedure of the PointWorld repository exactly. On the anchor frame of
each external view we place a $64\times64$ grid of query points and track them through
all $41$ frames with CoTracker3~\cite{karaev2024cotracker3}, the tracker used by
PointWorld's authors, at the native $192\times320$ resolution, which matches the depth
resolution and avoids rescaling. At every frame each tracked pixel is
lifted to 3D by sampling the ground-truth $S^2M^2$ depth of that frame and
unprojecting it with the camera intrinsics and the calibrated camera-to-world pose of
DROID-3D. Points with invalid depth ($z \le 0$ or $z > 4$~m) are discarded. We then
keep only points inside the robot workspace, a box of
$[0, 0.7]\times[-0.4, 0.4]\times[-0.3, 1.2]$~m in the robot base frame; this removes
floor, wall, and background tracks, which otherwise dominate the error. Remaining
spatial outliers, mostly tracks that jump between depth layers, are removed with a
multi-scale DBSCAN~\cite{ester1996dbscan} filter at radii $0.2$, $0.5$, and $1.0$~m,
dropping any track that is an outlier in a large fraction of frames. A point is
labelled \emph{moving} if its ground-truth displacement from the anchor frame exceeds
$3$~cm. Points on the robot arm,
identified by the URDF silhouette dilated by $3$~px, are excluded from all scored sets,
for the reasons given next.

\paragraph{Why robot points are excluded.}
In PointWorld the robot is an input, not a prediction target: the arm enters as points
computed from the joint states through the URDF (the gripper only, in the released
checkpoint), and its training clouds are built with the robot cut out. Seeding it with
arm pixels therefore asks it to predict the arm as if it were scenery, a case it never
saw in training, and the arm disintegrates within a few frames. Excluding the robot also
matches PointWorld's own evaluation, whose scene flows exclude it by construction. More
fundamentally, both models are conditioned on the robot's actions, so for PointWorld the
arm's future position is simply forward kinematics of the given joints, and scoring arm
points would test calibration rather than world modelling; the question worth comparing
is how the scene responds, i.e.\ objects, cloth, and doors. Finally, the arm would
otherwise dominate the metric: with the robot kept, about $178$ points per window
qualify as moving, against about $45$ without it, so roughly three quarters of a
robot-inclusive moving score would be arm motion that both models are handed. The
exclusion has costs. DepthWorld receives no credit for rendering the arm, which a video
model must do correctly. Because the mask is the dilated URDF silhouette, points on a
grasped object right at the gripper are dropped as well, as in PointWorld's own
pipeline. Cables remain in the scene, since they move with the arm but are not part of
the URDF.

\paragraph{Persistent point set.}
Tracked points become occluded over the horizon. Scoring whatever is valid at each
frame shrinks the scored set over time and biases it toward easy, unoccluded points,
which makes the error appear to \emph{decrease} with horizon. We therefore score a
fixed \emph{persistent} set per window: the moving points that are valid in every
frame of the horizon and that could be matched to a PointWorld seed (next paragraph),
about $45$ points per window. Error then increases with horizon, as it should. The
set is defined from the ground truth and PointWorld only, never from DepthWorld's
predictions, so PointWorld is a fixed reference when different DepthWorld checkpoints
are compared.

\paragraph{Prediction clouds.}
\emph{PointWorld.} PointWorld was trained with up to $12{,}000$ scene points, and
seeding it with only the few hundred tracked points is out of distribution and
understates it. We therefore seed it in
distribution with $12{,}000$ scene points sampled from the workspace-filtered,
robot-masked anchor-frame point map of both external views, supply the robot as URDF
points from the joint states as in its training, roll out the dense cloud with the
released large DROID checkpoint, and read off the trajectory of
each tracked point from its nearest dense seed (nearest-neighbour match on the anchor
frame, accepted if closer than $1$~cm). PointWorld thus receives full multi-view
context but is scored on exactly the tracked points.
\emph{DepthWorld.} For each tracked pixel and frame we bilinearly sample the predicted
depth of the corresponding external view, de-normalised to metric depth as in
Appendix~\ref{app:depth_decode}, and unproject it with the same intrinsics and pose
used for the ground truth.

\paragraph{Depth references and the own-ceiling comparison.}
For every window three depth stacks are available: the \emph{raw} ground-truth
$S^2M^2$ depth, the same depth after an encode--decode round trip through the SVD VAE
(\emph{VAE-GT}, the reference of Appendix~\ref{app:depth_decode}), and DepthWorld's
\emph{prediction}. The round trip is not lossless: on moving points it alone displaces
the ground truth by about $45$~mm (median), and because DepthWorld must operate in the
VAE's latent space, this is a floor its network cannot go below. Scoring DepthWorld
against raw depth therefore charges it for the VAE
floor, whereas scoring PointWorld against VAE-GT would penalise it for a smear it
never produced. We compute the full $2\times2$ of
\{PointWorld, DepthWorld\}$\times$\{raw, VAE-GT\} and report the diagonal as the
fair comparison: PointWorld against raw depth, and DepthWorld against VAE-GT. Each
model is thus measured against the best target its own representation can reach. The
off-diagonal entries are unfair in opposite directions and reverse the winner
depending on which is chosen. As a sanity check, the raw depth stored with
DepthWorld's outputs lifts to exactly the same 3D points as our own ground-truth
pipeline, ruling out any scale or alignment mismatch between the two pipelines.

\begin{figure}[!t]
    \centering
    \includegraphics[width=\linewidth]{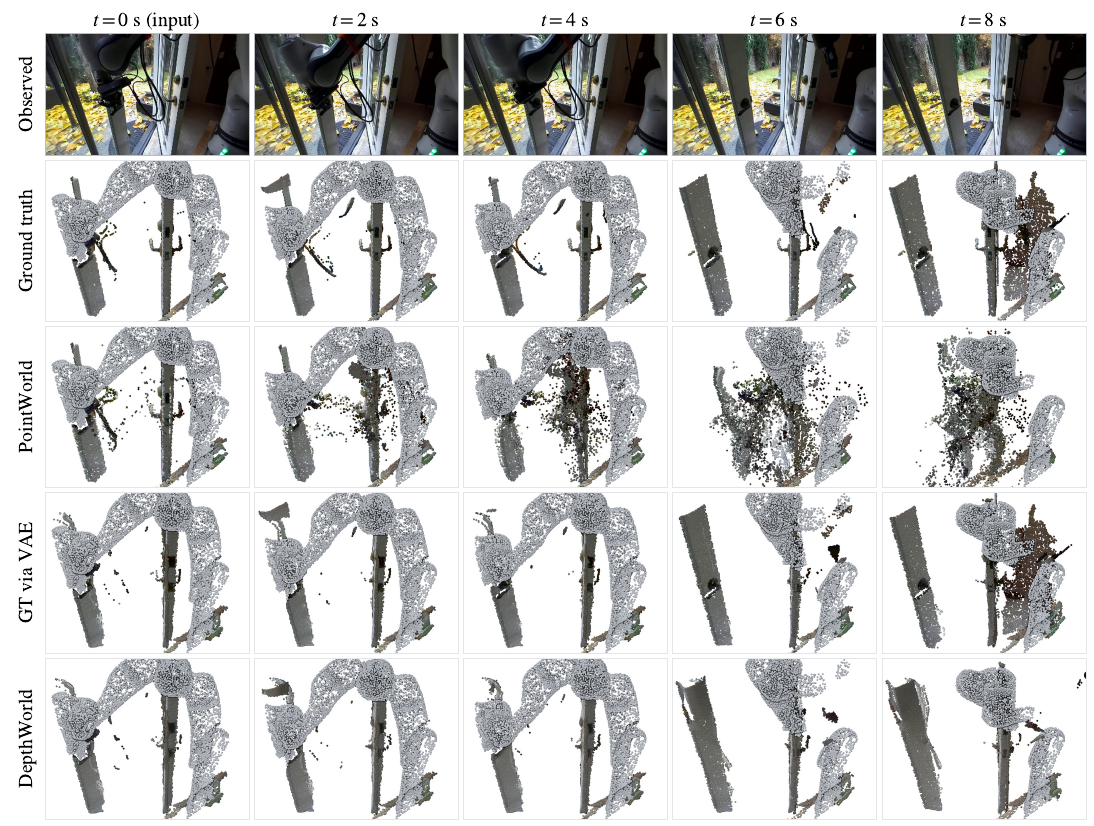}
    \caption{\textbf{Qualitative comparison with PointWorld} on a held-out window in which the
    robot opens a door. \emph{Top:} observed external RGB at the input frame and at $2$, $4$,
    $6$, and $8$~s. \emph{Rows 2--5:} workspace point clouds in the robot frame at the same
    times: the ground truth lifted from $S^2M^2$ depth; PointWorld's dense rollout seeded at
    $t=0$; the ground truth after the SVD-VAE round trip, which is DepthWorld's own reference
    in Table~\ref{tab:pointworld_cmp}; and DepthWorld's predicted depth unprojected into the
    robot frame. PointWorld follows the arm at short horizon, but from about $4$~s its cloud
    scatters and drifts, consistent with its rising whole-scene Chamfer distance. DepthWorld
    stays coherent over the full $8$~s and reconstructs the opening door in agreement with
    its VAE reference, at the cost of thin structures and some foreground geometry that are
    present in the raw ground truth.}
    \label{fig:pw_qual}
\end{figure}

\begin{table}[!t]
\centering \small
\setlength{\tabcolsep}{5pt}\renewcommand{\arraystretch}{1.15}
\caption{\textbf{Comparison with PointWorld} on the shared $256$-window split ($141$
windows for the moving-point metrics, $147$ for the whole scene), fair diagonal of the
own-ceiling protocol: PointWorld is scored against raw ground-truth depth, DepthWorld
(the RGB+D PM-DPT variant) against the VAE round-tripped ground truth. Medians over
windows in mm at horizons of $2$~s and $8$~s; $\ell_2$ is averaged over the frames up
to the horizon, Chamfer is evaluated at the horizon frame. Lower is better; best per
column in \textbf{bold}.}
\label{tab:pointworld_cmp}
\begin{tabular}{l cc cc cc}
\toprule
 & \multicolumn{2}{c}{Tracked-point $\ell_2$, moving} & \multicolumn{2}{c}{Chamfer, moving} & \multicolumn{2}{c}{Chamfer, whole scene} \\
\cmidrule(lr){2-3}\cmidrule(lr){4-5}\cmidrule(lr){6-7}
Model & 2\,s & 8\,s & 2\,s & 8\,s & 2\,s & 8\,s \\
\midrule
PointWorld~\cite{huang2026pointworld} vs.\ raw GT & \textbf{23} & 56 & \textbf{21} & 58 & \textbf{8} & 18 \\
DepthWorld vs.\ VAE-GT & 31 & \textbf{46} & 25 & \textbf{32} & \textbf{8} & \textbf{9} \\
\bottomrule
\end{tabular}
\end{table}

\paragraph{Metrics and scoring regions.}
We report two metrics in two regions, as medians over windows; the $\ell_2$ error is
averaged over the frames up to the horizon and the Chamfer distance is evaluated at
the horizon frame. The \emph{tracked-point $\ell_2$} error is the
Euclidean distance between the predicted and target 3D position of the same point,
averaged over the persistent moving set; it penalises a point that lands on the wrong
part of the scene and is only possible because the persistent set provides stable
correspondences. The \emph{symmetric Chamfer} distance,
$\tfrac12\bigl(\mathrm{mean}\,\mathrm{NN}(\text{pred}\to\text{target}) +
\mathrm{mean}\,\mathrm{NN}(\text{target}\to\text{pred})\bigr)$, is
correspondence-free and measures whether the predicted cloud occupies the right
space, regardless of point identity. The \emph{moving region} scores only moving
points and tests whether the model predicts the manipulation itself. The \emph{whole
scene} scores all workspace points, subsampled to $8{,}000$ per cloud for equal
density, and is dominated by static structure: it mostly measures whether the scene
is reconstructed without drift. A Chamfer distance much smaller than the $\ell_2$
error indicates that the predicted cloud has roughly the right shape but that
correspondences have drifted.

\paragraph{Results.}
Table~\ref{tab:pointworld_cmp} reports the fair diagonal at $2$~s and $8$~s, and
Figure~\ref{fig:pw_qual} shows a representative window.
PointWorld is more accurate on tracked moving points at short horizon ($23$ vs.\
$31$~mm at $2$~s) but degrades faster, and DepthWorld is ahead at $8$~s ($46$ vs.\
$56$~mm). On the whole scene the two are equal at $2$~s ($8$~mm), after which
DepthWorld's Chamfer distance stays flat ($9$~mm at $8$~s) while PointWorld's grows to
$18$~mm: the depth-image representation reconstructs the static scene without drift,
whereas PointWorld's point cloud drifts as the rollout proceeds. In the moving region
PointWorld is slightly ahead at $2$~s ($21$ vs.\ $25$~mm) and DepthWorld degrades far
less by $8$~s ($32$ vs.\ $58$~mm). The DepthWorld model in this comparison is the RGB+D PM-DPT variant of
Table~\ref{tab:wm_quality}. We stress the asymmetries noted above: DepthWorld answers an easier query
and is scored above its VAE floor. Against raw depth, PointWorld's point
representation retains a large advantage on moving objects, a representational cost
of latent-space depth prediction that we consider an important direction for future
work.

\subsection{Comparison with TesserAct (Supplement to Section~\ref{sec:eval})}
\label{app:tesseract_cmp}
TesserAct~\cite{zhen2025tesseract} is a text-conditioned, single-view RGB--depth--normal
(RGBDN) video model built on CogVideoX-5b-I2V~\cite{yang2024cogvideox}, a 5B-parameter
image-to-video diffusion transformer. It is not comparable off the shelf: it has no
action conditioning, sees a single camera, and was trained on different data. We
therefore adapt it minimally to our setting and retrain it on the same episodes as
DepthWorld at an equal step budget.

\paragraph{Model and adaptation.}
The TesserAct architecture is kept unchanged. Its patch embedding splits the latent into
three modality streams (RGB, depth, normal), projects each, and sums them into one shared
token sequence, and two additional output heads produce the depth and normal latents.
The
transformer is initialised from the public CogVideoX-5b-I2V weights rather than from
TesserAct's released RGBDN checkpoint, mirroring DepthWorld, which starts from the
generic SVD checkpoint rather than from a depth-pretrained model. The single
conceptual change is the conditioning signal. CogVideoX conditions on T5 text tokens
passed as the cross-modal context of its joint-attention blocks. We replace these with
action tokens: for each of the $9$ frames of a clip, the 7-D end-effector pose and
gripper state is mapped by a small MLP ($7 \to 1024 \to 4096$ with a SiLU
non-linearity), added to a learned per-frame temporal embedding, and layer-normalised,
giving $9$ tokens of width $4096$ that occupy the T5 slot. Because CogVideoX-5b-I2V uses
rotary position embeddings, the context length is free, so $9$ action tokens replace the
$226$ padded text tokens without touching any transformer block. This mirrors
DepthWorld's per-frame end-effector action conditioning. For classifier-free guidance,
the whole trajectory is replaced by an all-zero null trajectory with probability
$0.05$, and the conditioning frame is dropped with the same probability.

\paragraph{Data.}
We use the same DROID-3D training episodes as DepthWorld, restricted to the two external
cameras; the wrist camera is excluded, and each external camera is treated as an
independent single-view sample, since TesserAct has no notion of a multi-view rig.
Clips are contiguous $9$-frame windows at $5$~Hz and $192\times320$ resolution, with the
per-frame actions aligned to the same timestamps. Depth is the DROID-3D metric depth
mapped to TesserAct's log-grayscale encoding over $0.30$--$3.30$~m and replicated to
three channels; normals are computed geometrically from the depth; actions are
percentile-normalised per dimension. This yields $144{,}436$ training clips and
$1{,}534$ validation clips.

\paragraph{Training.}
The training objective is TesserAct's own: given the VAE-encoded first RGBDN frame and
the action trajectory, denoise the $9$-frame RGBDN clip with CogVideoX's velocity
target, with the loss being the sum of three separately weighted per-modality MSE terms.
Each modality is encoded separately by the frozen CogVideoX VAE. Both the transformer
and the action encoder are trained. We train on $16$ H200 GPUs with data-parallel
training in bf16 with gradient checkpointing, full-precision AdamW ($\beta_1 = 0.9$,
$\beta_2 = 0.95$, weight decay $10^{-3}$, gradient clipping at $1.0$), a global batch
of $64$ clips, and a constant learning rate of $5\times10^{-5}$ after a $200$-step
warm-up. We train for $50$k steps (about $20$ hours), matching the $50$k-step budget of
the DepthWorld model it is compared against.

\paragraph{Evaluation.}
We mirror the DepthWorld protocol of Appendix~\ref{app:wm_eval}. Generation uses the
released TesserAct pipeline without its text encoder: the action tokens are passed as
the prompt embedding and the null trajectory as the negative embedding, with
classifier-free guidance scale $6.0$ and $50$ DPM-Solver steps at $192\times320$.
Rollouts are autoregressive and closed-loop: five chained $9$-frame generations, each
re-anchored on the last generated RGBDN frame (ground truth is never re-injected) and
given the next $8$-frame action segment, producing $1 + 5\times8 = 41$ frames or
$8.2$~s, the same horizon as DepthWorld, which reaches it in ten steps of four new frames.
The
evaluation set is $128$ held-out samples, each an episode with a fixed anchor frame
scored on both external views; the two views are averaged with equal weight. As for DepthWorld, metrics are computed against the ground truth
after an encode--decode round trip through the model's own VAE (here the CogVideoX
VAE), so that each model is scored above its own representational floor
(Appendix~\ref{app:depth_decode}); we additionally record the error against raw depth
and the VAE floor itself. Metric depth is recovered by inverting the log encoding, the
conditioning frame is excluded, and the RGB and depth metrics of
Appendix~\ref{app:wm_eval} are averaged over the $40$ generated frames.

\paragraph{Results.}
At the same $50$k steps, averaged over the $40$ generated frames on the external views,
the adapted TesserAct reaches $19.43$~dB PSNR and $0.154$ AbsRel, against $24.07$~dB
and $0.074$ for the DepthWorld RGB+D model (the S$^2$M$^2$ row of
Table~\ref{tab:depth_source}): it trails by $4.6$~dB and about $2\times$ in AbsRel.
The gap is mostly one of rollout
stability rather than single-step quality: TesserAct's first generated chunk is
reasonable (PSNR $23.55$, LPIPS $0.065$, AbsRel $0.082$, $\delta_1$ $0.95$ over its
eight future frames), but quality decays steadily along the rollout, from PSNR $27.9$,
AbsRel $0.048$, and $\delta_1$ $0.97$ at the first generated frame to $16.8$, $0.211$,
and $0.81$ at the fortieth, with no visible seams at the re-anchoring boundaries.
Qualitatively, the generated arm follows the commanded trajectory over the full $8$~s,
indicating that the action conditioning is used. Two differences should be kept in mind when reading these numbers. TesserAct sees a single view and so has no cross-view
context, whereas DepthWorld predicts all cameras jointly. Its depth also passes through
the CogVideoX VAE, whose floor on raw depth is large (AbsRel $0.28$ on one external
camera), so its raw-depth error is dominated by the representation rather than the
network; scoring both models above their own VAE floor, as we do, removes this effect
from the comparison. The guidance scale was not tuned.


\end{document}